\pdfoutput=1

\documentclass[11pt]{article}
\usepackage{hyperref}
\usepackage{url}

\usepackage{graphicx}
\usepackage{listings}
\usepackage{amsthm}
\usepackage{csquotes}
\usepackage{verbatim}
\usepackage{subcaption}
\usepackage{multirow,multicol}
\usepackage{amsmath,amsfonts,amssymb,amsthm}
\theoremstyle{definition}

\usepackage[linesnumbered,ruled,vlined]{algorithm2e}
\SetKwInput{KwInput}{Input}
\SetKwInput{KwOutput}{Output}

\usepackage[]{EACL2023}

\usepackage{times}
\usepackage{latexsym}

\usepackage[T1]{fontenc}

\usepackage[utf8]{inputenc}

\usepackage{microtype}

\usepackage{inconsolata}
\usepackage{graphicx}
\usepackage{caption}
\usepackage{subcaption}
\title{How Many and Which Training Points Would Need to be Removed \\ to Flip this Prediction?}

\author{Jinghan Yang \\
  The University of Hong Kong \\
  and Northeastern University\\
  \text{eciel@connect.hku.hk} \\
  \And
  Sarthak Jain \\
  AWS AI Labs\thanks{~~Work done prior to joining Amazon.} \\
  \text{jsarth@amazon.com} \\
  \And
  Byron C. Wallace \\
  Northeastern University \\
  \text{b.wallace@northeastern.edu} \\}

\begin{document}
\maketitle
\begin{abstract}

We consider the problem of identifying a \emph{minimal subset} of training data $\mathcal{S}_t$ such that if the instances comprising $\mathcal{S}_t$ had been removed prior to training, the categorization of a given test point $x_t$ would have been different.
Identifying such a set may be of interest for a few reasons.
First, the cardinality of $\mathcal{S}_t$ provides a measure of robustness (if $|\mathcal{S}_t|$ is small for $x_t$, we might be less confident in the corresponding prediction), which we show is correlated with but complementary to predicted probabilities.
Second, interrogation of $\mathcal{S}_t$ may provide a novel mechanism for \emph{contesting} a particular model prediction: If one can make the case that the points in $\mathcal{S}_t$ are wrongly labeled or irrelevant, this may argue for overturning the associated prediction. 
Identifying $\mathcal{S}_t$ via brute-force is intractable.
We propose comparatively fast approximation methods to find $\mathcal{S}_t$ based on \emph{influence functions}, and find that---for simple convex text classification models---these approaches can often successfully identify relatively small sets of training examples which, if removed, would flip the prediction.\footnote{Code and data to reproduce all experiments available at: \url{https://github.com/ecielyang/Smallest_set}}

\end{abstract}

\section{Introduction}

In this work we pose the following problem in the context of binary classification: \emph{For a test point $x_t$, identify a minimum subset $\mathcal{S}_t$ of training data that one would need to remove in order to flip the prediction $\hat{y}_t$ for $x_t$}. 
This subset may be of interest for a few reasons.
First, the cardinality $k$ of $\mathcal{S}_t$ captures one measure of the (in)fragility of the prediction $\hat{y}_t$: 
Small $k$ indicates that a minor change in the training data would have resulted in a different (discrete) prediction for $x_t$. 
We later show that this measure is correlated with, but complementary to, predicted probabilities. 

Perhaps a more interesting motivation for recovering $\mathcal{S}_t$ is to provide a potential mechanism for \emph{contesting} model predictions \citep{hirsch2017designing,vaccaro2019contestability}, i.e., to enable individuals to interrogate and dispute automatic determinations that affect them.
If removing a small set of training points would have yielded a different prediction, and if one could make the case for excluding these points (e.g., because they seem mislabeled, or reflect systematic labeling biases), this might provide a compelling case to overturn a model prediction. 
Consider an educator using an automated essay grading system.\footnote{We put aside the question of whether using automated approaches in this particular setting is appropriate to begin with (likely not, though this may depend on how it is used).}
Assume the system has output a comparatively poor grade for a student, a determination they see as unfair.
Contesting the inclusion of a small set of examples ($\mathcal{S}_t$) which, if excluded, would have resulted in a higher grade provides a novel mechanism for disputation. 

Naïvely attempting to find $\mathcal{S}_t$ by brute enumeration and re-training would be hopelessly inefficient. 
We introduce an algorithm for finding such sets efficiently using \emph{influence functions} \cite{IF} which allow us to approximate changes in predictions expected as a result of removing subsets of training data \cite{koh2019accuracy}. 
We then provide an iterative variant of this method which does a better job of identifying sets $\mathcal{S}_t$. 

Across different datasets and models, we find that we are often able to recover subsets $\mathcal{S}_t$ with relatively small cardinality $k$; i.e., one can often identify a small to medium subset of training data which, if removed, would flip a given prediction.
We also find that there are many test points for which models make predictions with high confidence but where $k$ is small.

The {\bf contributions} here include an investigation of the task of identifying minimal training sets to flip particular predictions in the context of text classification, algorithms for this problem, and an empirical evaluation of their performance in the context of binary text classification.\footnote{Recent related work in economics by \citet{broderick2020automatic} proposed and investigated a similar problem, with a focus on identifying the sensitivity of econometric analyses to removal of small subsets of data. Recent work on \emph{datamodeling} \cite{pmlr-v162-ilyas22a} also considered a variant of this problem (see Section \ref{section:related-work}).}

\section{Methods}
\label{section:methods} 

Assume a binary text classification problem. 
Given a training set $Z^{\text{tr}} = {z_1, . . . , z_N}$, where $z_i = (x_i, y_i) \in \mathcal{X} \times \mathcal{Y}$,  
we 
aim to estimate the parameters $\theta$ of a classification model $f_{\theta} : \mathcal{X} \rightarrow \mathcal{Y}$ to minimize the empirical risk, i.e., loss $\mathcal{L}$ over $Z^{\text{tr}}$: $\hat{\theta} := 
{\text{argmin}}_{\theta} \frac{1}{N}\sum_{i=1}^N \mathcal{L}(z_i, \theta)  + \frac{\lambda}{2} \theta^T \theta$, 
which we will denote by $R(\theta)$. 
We assume throughout that $R$ is twice-differentiable and strongly convex in $\theta$, i.e.,
$H_{\hat{\theta}} := \nabla_{\theta}^2 R(\hat{\theta}) 
:= \frac{1}{N} \sum_{i=1}^N \nabla_{\theta}^2  \mathcal{L}(z_i, \hat{\theta}) + {\lambda} \mathit{I} $
exists and is positive definite. 
Suppose we removed a subset of $k$ training points $\mathcal{S} \subset Z^{\text{tr}}$ and re-estimated $\theta$, 
yielding 
new parameters $\hat{\theta}_{\mathcal{S}}$.
Letting $\varepsilon = -\frac{1}{N}$, we can write this as:

\begin{equation}
\hat{\theta}_{S}={\text{argmin}}_{\theta \in \Theta} \{R(\theta) + \varepsilon \sum_{z_i \in S} \mathcal{L}(z_i, \theta)\}
\end{equation}

In principle, one could remove the points in $\mathcal{S}$ and re-train 
to find $\hat{\theta}_{\mathcal{S}}$. 
In practice this is infeasible given the number of potential subsets $\mathcal{S}$. 
\citet{IF} provide (relatively) efficient approximations to estimate $\hat{\theta}_{\mathcal{S}}$ when $k$=1. 
Subsequent work \citep{koh2019accuracy} found that this approximation correlates well with the actual empirical effects of removing a \emph{set} of points (where $k>1$).




\vspace{0.4em}
\noindent{\bf Finding influential subsets} Given input $x_t$, we aim to design an approach to efficiently \emph{identify} the smallest subset $\mathcal{S}_t$ of $Z^{\text{tr}}$ such that removing these examples prior to training would change $\hat{y}_t$. 
Prior work \citep{cook,IF} derived the influence exerted by a train point $i$ on the \emph{loss} incurred for a test point $t$ as: 

\begin{equation}
-\nabla_\theta \mathcal{L}(z_t, \hat{\theta})^\intercal \underbrace{H^{-1}_{\hat{\theta}} \nabla_{\theta} \mathcal{L}(z_i, \hat{\theta})}_{\Delta_i \theta}
\label{eq:inf-classic-loss}
\end{equation}

\noindent Where $\Delta_i \theta$ is the influence of upweighting $z_i$ during training on estimates $\hat{\theta}$ \citep{cook}.
We are interested, however, in identifying points that have a particularly strong effect on a specific observed \emph{prediction}.
We therefore modify Equation \ref{eq:inf-classic-loss} to estimate the \emph{influence on prediction} (IP), i.e., the change in predicted probability 
for $x_t$ observed after removing training instance $i$. This can be expressed as:


\begin{equation}
    \Delta_{t}f_i := - \nabla_{\theta} f_{\hat{\theta}}(x_t)^\intercal \Delta_i \theta
    \label{eq:inf-IP}
\end{equation}


\noindent We then approximate the change in prediction on instance $t$ we would anticipate after removing the training subset $\mathcal{S}_t$ from the training data as the sum of the $\Delta_{t}f_i$ terms for all points $x_i \in \mathcal{S}_t$. 



Algorithm \ref{alg:alg1} describes a method for constructing $\mathcal{S}_t$. 
We estimate the change in output expected upon removing each instance from the training dataset and assemble these in $\Delta_t f$.
We then greedily consider adding these differences (effectively adding points to $\mathcal{S}_t$) until the resultant output is expected to cross the classification threshold ($\tau$); if we exhaust the training dataset without crossing $\tau$, then we have failed to identify a set $\mathcal{S}_t$. 


\begin{algorithm}
\caption{A simple method to find a minimal subset to flip a test prediction}
	\label{alg:alg1}
\DontPrintSemicolon
  \KwInput{$f$: Model;
  $Z^{\text{tr}}$: Full training set; 
  $\hat{\theta}$: Parameters estimated $Z^{\text{tr}}$;  
  $\mathcal{L}$: Loss function; 
  $x_{t}$: A test point; 
  $\tau$: Classification threshold (e.g., 0.5)}
  \KwOutput{$\mathcal{S}_t$: minimal train subset identified to flip the prediction ($\emptyset$ if unsuccessful)}
  $H \leftarrow \nabla_{\theta}^2  \mathcal{L}(Z^{\text{tr}}, \hat{\theta})$ \\ 
  $\Delta \theta \leftarrow H^{-1} \nabla_{\hat{\theta}} \mathcal{L}(Z^{\text{tr}}, \theta')$ \\ 
  $\Delta_t f \leftarrow \nabla_{\theta} f_{\hat{\theta}}(x_{t})^\intercal \Delta \theta$\\
  $\hat{y}_t \leftarrow f(x_{t}) > \tau$ \tcp*{Binary prediction}
    \tcp{Sort instances (and estimated output differences) in order of the current prediction} 
  ${\tt direction} \leftarrow \{\uparrow$ if $\hat{y}_t$ else $\downarrow$\} \\
  ${\tt indices} \leftarrow {\tt argsort}(\Delta_t f, {\tt direction})$\\
  $\Delta_t f \leftarrow {\text{\tt sort}}(\Delta_t f, {\tt direction})$ \\
   	\For{$k=1$ ... $|Z^{\text{tr}}|$} 
   	{
   	$\hat{y}_t' = (f(x_{t}) + {\text{\tt sum}}(\Delta_t f [:k])) > \tau$ 
   	
   	\If{$\hat{y}'_t \neq \hat{y}_t$}
        {
            \KwRet $Z^{\text{tr}}[{\tt indices}[:k]]$
        }
   	}
   	\KwRet $\emptyset$
\end{algorithm}

Algorithm \ref{alg:alg1} is simple and relatively fast, but we can improve upon it by \emph{iteratively} identifying smaller subsets $\mathcal{S}_t$ in Algorithm \ref{alg:iterative}. 
We detail this approach in Appendix Algorithm \ref{alg:iterative}, but describe it briefly as follows.

We start with the entire train set as a ``candidate'' $\mathcal{\tilde{S}}_t$, and then iteratively attempt to find strict subsets of this that by themselves would flip the prediction $\hat{y}_t$. 
On the first pass, this is equivalent to Algorithm \ref{alg:alg1}, after which---if successful---we will have found a candidate set $\mathcal{\tilde{S}}_t$.
Here we update parameter estimates $\theta$ to approximate ``removing'' the points in $\mathcal{\tilde{S}}_t$, and then we recompute the approximation of the influence that points in $\mathcal{\tilde{S}}_t$ would have on $\hat{y}_t$ using a single-step Newton approximation. 
The idea is that after the parameter update this approximation will be more accurate, potentially allowing us to find a smaller $\mathcal{S}_t$. 
This process continues until we are unable to find a new (smaller) subset.

In sum, this variant of the algorithm attempts to iteratively identify increasingly small subsets $\tilde{\mathcal{S}}_t$ which would, upon removal prior to training, overturn the original prediction $\hat{y}_t$. 
There is a computational cost to this, because each iteration involves approximating the influence on a particular prediction; this is computationally expensive. 
This variant therefore trades run-time for (hopefully) more accurate identification of minimal $\mathcal{S}_t$.
However, we find that empirically Algorithm \ref{alg:iterative} ends up running for only $2.3$ passes on average (across all experiments).
That is, the algorithm adds a scalar to the run-time of Algorithm \ref{alg:alg1}, but often yields considerably smaller $\mathcal{S}_t$. We show the comparison of $|\mathcal{S}_t|$ returned by two algorithms in the Table \ref{table:k-compare}.

\begin{table}
\small
\centering 
\begin{tabular}{l l l }
\hline
{\textbf{Models}} & {\textbf{Algorithm \ref{alg:alg1}}} & {\textbf{Algorithm \ref{alg:iterative}}} \\
\hline 
 \multicolumn{3}{c}{\emph{Movie reviews}} \\
BoW & 247 & 151  \\
BERT & 484 & 303\\
\multicolumn{3}{c}{\emph{Essays}} \\
BoW      & 352 & 134\\
BERT & 484 & 135 \\
 \multicolumn{3}{c}{\emph{Emotion classification}} \\
BoW  & 500 & 345 \\
BERT & 524 & 327 \\
 \multicolumn{3}{c}{\emph{Hate speech}} \\
BoW      & 808 & 415\\
BERT & 546 & 239\\
\multicolumn{3}{c}{\emph{Tweet sentiment}} \\
BoW         & 345 & 177\\
BERT        & 858 & 569\\
\hline
\end{tabular}
\caption{The comparison of average on $k = |\mathcal{S}_t|$ values from Algorithm \ref{alg:alg1} and Algorithm \ref{alg:iterative} over the subsets of test points $x_t$ for which we were able
to successfully identify a set of points $|\mathcal{S}_t|$ }
\label{table:k-compare}
\end{table}






\section{Experimental Setup}
\label{section:experiments}

\begin{table}
\small
\centering 
\begin{tabular}{l l l}
\hline
{\textbf{Features}} & {\textbf{Found $\mathcal{S}_t$}} & {\textbf{Flip successful}} \\
\hline 
 \multicolumn{3}{c}{\emph{Movie reviews}} \\
BoW & 78\% & 78\% \\
BERT & 79\% &  72\% \\ 
\multicolumn{3}{c}{\emph{Essays}} \\
BoW          &  12\% & 11\% \\ 
BERT &  9\% & 8\%\\
 \multicolumn{3}{c}{\emph{Emotion classification}} \\
BoW          & 91\% & 91\% \\ 
BERT & 83\% & 71\% \\ 
 \multicolumn{3}{c}{\emph{Hate speech}} \\
BoW          &  67\% & 60\%    \\ 
BERT &  53\% & 44\%     \\
\multicolumn{3}{c}{\emph{Tweet sentiment}} \\
BoW         & 99\% & 91\%  \\
BERT        & 90\% & 68\%   \\
\hline
\end{tabular}
\caption{Percentages of test examples for which Algorithm \ref{alg:iterative} successfully identified a set $\mathcal{S}_t$ to remove (center) and for which upon removing these instances and retraining the prediction indeed flipped (right).}
\label{table:k-flips}
\end{table}

\noindent{\bf Datasets} We use five binary text classification tasks: Movie review sentiment 
\cite{socher2013recursive}; Twitter sentiment classification \cite{go2009twitter}; Essay grading \cite{essay_data}; Emotion classification \cite{saravia-etal-2018-carer}, and; Hate speech detection \cite{gibert2018hate}. 
We binarize the essay data by labeling the top 10\% score points as 1 (``A''s) and others as 0. 
For the emotion dataset, we include only ``joy'' and ``sadness''.  
We provide dataset statistics in Appendix Table \ref{table:dataset-info}.
Because the hate speech data is severely imbalanced, we selected a classification threshold $\tau$ post-hoc in this case to maximize train set F1 (yielding $\tau=0.25$); for other datasets we used $\tau=0.5$, which corresponded to reasonable F1 scores---for reference we report prediction performance on all datasets in  Appendix Table \ref{table:model-performance}.


\vspace{0.25em}
\noindent{\bf Models} We consider only $\ell$2 regularized logistic regression (for which influence approximation is well-behaved). 
As features, we consider both bag-of-words and neural embeddings (induced via BERT; \citealt{devlin2018bert}).


\section{Results}
\label{section:results} 
\begin{figure*}
\centering
\begin{subfigure}{0.49\textwidth}
    \includegraphics[width=\textwidth]{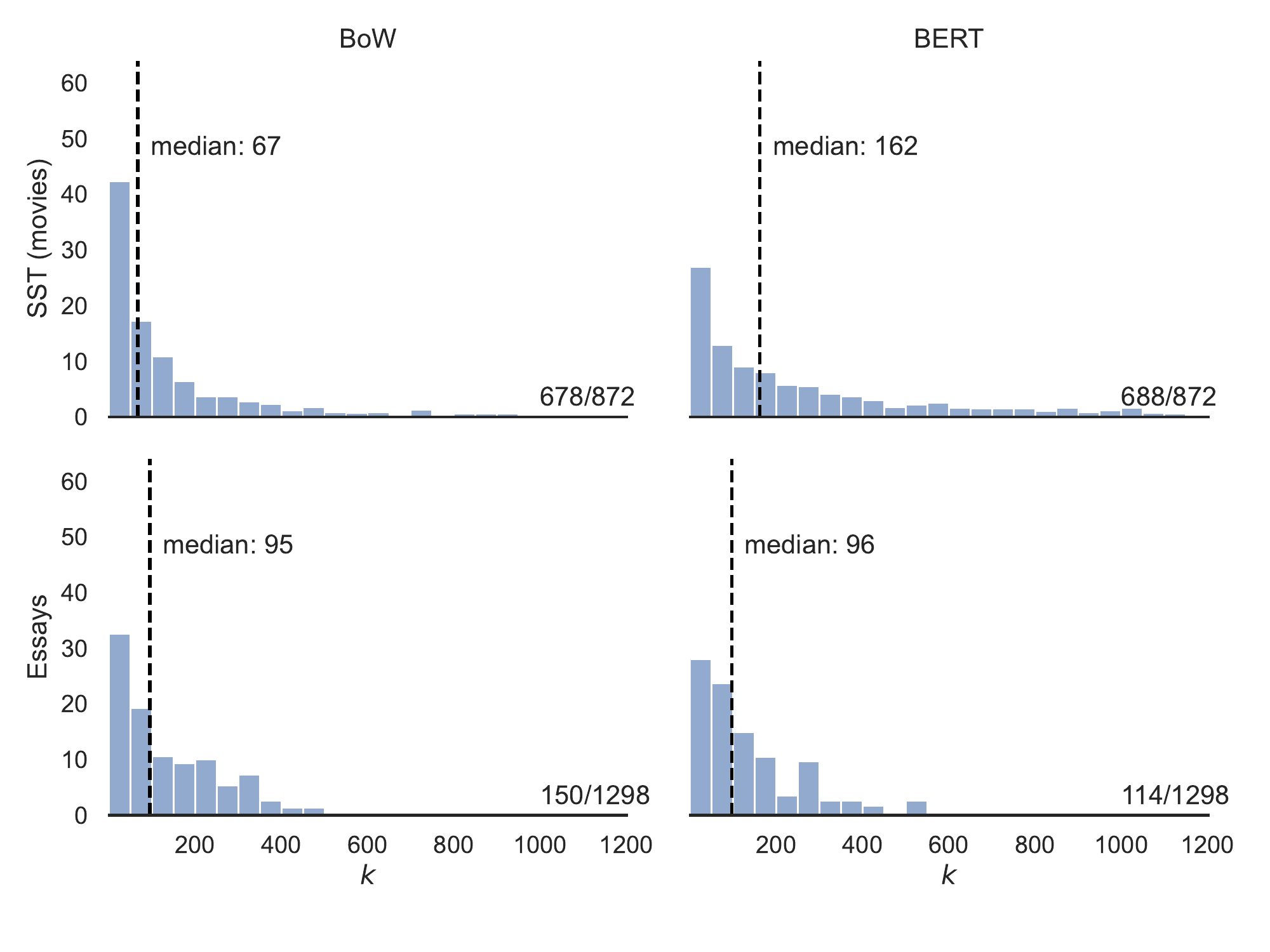}
    \caption{Histograms of $k=|\mathcal{S}_t|$ values over the subsets of test points $x_t$ for which we were able to successfully identify a set of points $\mathcal{S}_t$ such that removing them would flip the prediction for $\hat{y}_t$. We report the fraction for which we were able to do so in the lower sub-plot right corners.}
    \label{fig:k-movies-LR-alg4}
    \end{subfigure}
    ~
\begin{subfigure}{0.49\textwidth}
    \includegraphics[width=\linewidth]{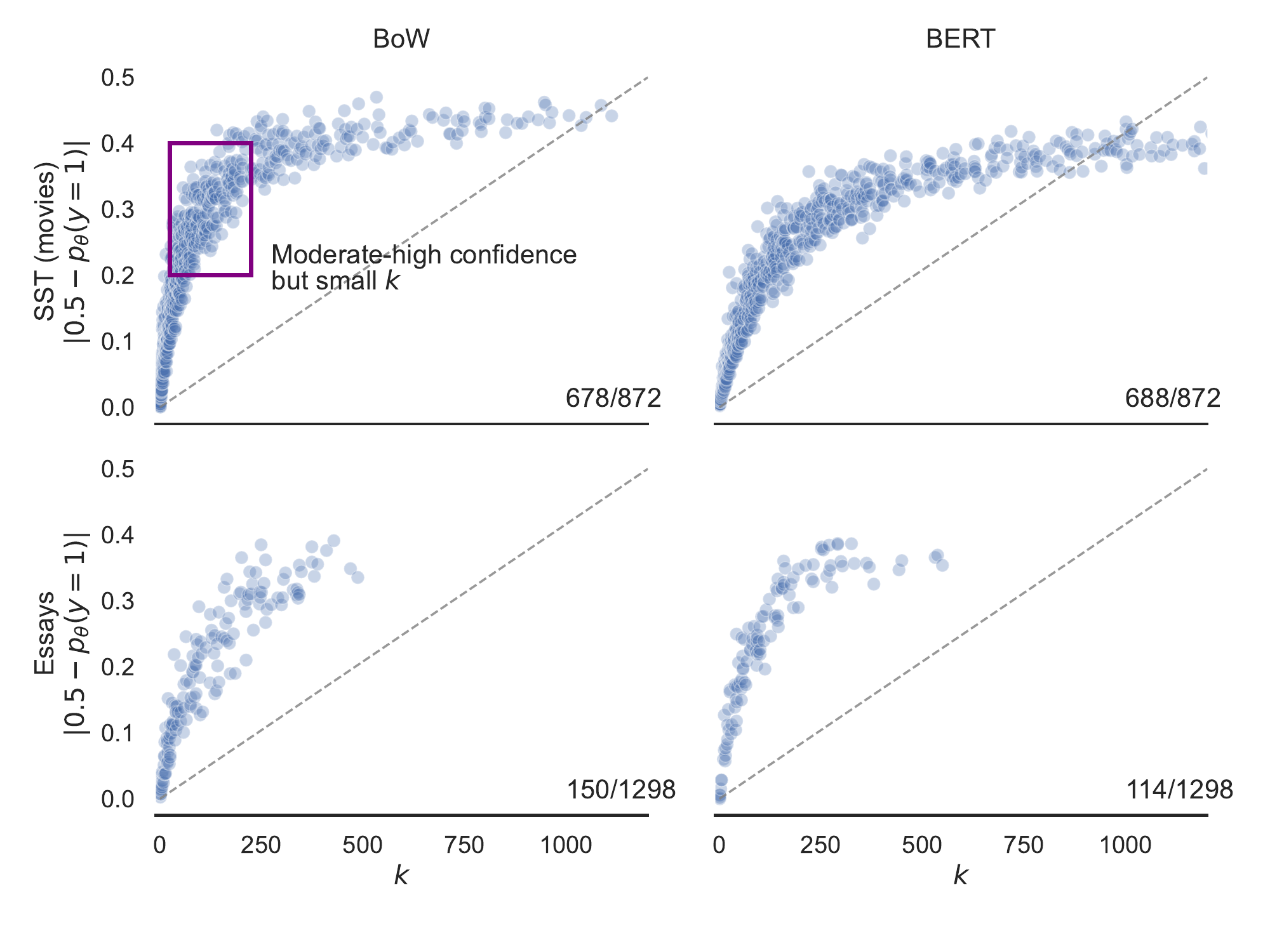}
    \caption{Relationship between predicted probabilities and  $k=|\mathcal{S}_t|$ identified. These are correlated (as we would expect), but there are many points for which the model is moderately or highly confident, but where removing a relatively small set of training data would change the prediction.} 
    \label{fig:k-v-p}
    \end{subfigure}
    \caption{Results characterizing $\mathcal{S}_t$ on two illustrative datasets (sentiment classification and essay scoring).} 
\end{figure*}

Here we present results for the iterative method (Appendix Algorithm \ref{alg:iterative}), which outperforms the simpler Algorithm \ref{alg:alg1}.
We provide full results for both methods in the in the Appendix. 

\vspace{0.5em}
\noindent{\bf How often can we find $\mathcal{S}_t$ and how frequently does removing the instances it contains flip the prediction?} 
As can be seen in Table \ref{table:k-flips}, this varies considerably across datasets.
For movie reviews, Algorithm \ref{alg:iterative} returns a set $\mathcal{S}_t$ for $\sim$80\% of test points, whereas for the (more complex) essays data it does so for only $\sim$10\% of instances. 
Other datasets see success somewhere in-between these extremes.
However, when the algorithm does return a set $\mathcal{S}_t$, removing this and re-training almost always flips the prediction $\hat{y}_t$ (right-most column). 

\vspace{0.5em}
\noindent{\bf What is the distribution of $k=|\mathcal{S}_t|$?} 
Figure \ref{fig:k-movies-LR-alg4} shows empirical distributions of $k$ values for subsets $\mathcal{S}_t$ identified by Algorithm \ref{alg:iterative} for the illustrative movie review and essay grading datasets (full results in Appendix).
The take-away here is that when we do find $\mathcal{S}_t$, its cardinality is often quite small. 
Indeed, for many test points removing tens of examples would have flipped the prediction.

\vspace{0.5em}
\noindent {\bf How does $k$ relate to predicted probability?}
Does the size of $\mathcal{S}_t$ tell us anything beyond what we might infer from the predicted probability $p(y_t=1)$? 
In Figure \ref{fig:k-v-p} we show (again for just two datasets here) a scatter of $k=|\mathcal{S}_t|$ against the distance of the predicted probability from 0.5. 
The former provides complementary information, in that there exist instances about which the model is confident, but where removing a small set of training instances would overturn the prediction.

\vspace{0.5em}
\noindent{\bf Qualitative example.} One reason to recover sets $\mathcal{S}_t$ is to support \emph{contestation}---if $k$ is small, one might argue against the appropriateness of the points in $\mathcal{S}_t$ and hence against the determination $y_t$.
As a simple example,\footnote{We provide more qualitative analysis in the Appendix.} consider the movie review test instance ``\emph{Manages to transcend the sex drugs and show tunes plot into something far richer}''. 
The true label is positive, but the model predicted negative. 
Algorithm \ref{alg:iterative} reveals that removing a single example ($k=1$) from the training set would have reversed the prediction---specifically, this negative review: ``\emph{An overstylized pureed melange of sex psychology drugs and philosophy}''.
It seems this training point is only superficially similar to the test point, which may make a case for overturning the prediction. 
While standard influence functions \cite{IF} can be used to \emph{rank} training points, the novelty here is observing that \emph{removing this point alone} would change the prediction.

\section{Related Work}
\label{section:related-work}

\vspace{0.5em}
\noindent{\bf Influence functions} \citep{hampel1974influence,cook1980characterizations,cook} provide machinery to identify training points that most informed a particular test prediction.
Influence can provide insight into predictions made by modern neural networks \citep{IF}, and can be used to \emph{debug} models and training data by surfacing mislabeled training points and/or reliance on artifacts   \citep{adebayo2020debugging, han2020explaining,pezeshkpour-etal-2022-combining,teso2021interactive}, and tuning influence can be used to demote reliance on unwanted correlations \cite{han-tsvetkov-2021-influence-tuning}. 
Influence can also be used to \emph{audit} models by inspecting training data responsible for predictions \citet{marx2019disentangling}.  

\citet{schulam2019can} audit individual predictions by approximating how much they might have changed under different samples from the training distribution. 
\citet{ting2018optimal} consider influence functions as a tool for optimally subsampling data in service of computational efficiency. 
\citet{koh2019accuracy} considered approximating the effect of removing a \emph{group} of training points using influence functions, and found that they do so fairly well (a result that we use). 
They assumed groups were \emph{given} and then evaluated the accuracy of the influence approximation to the change in prediction. 
By contrast, we are interested in \emph{finding} a (minimal) group which would have the specific effect of flipping a prediction. 
Elsewhere, \citet{khanna2019interpreting} ask: ``Which training examples are most responsible for a given \emph{set} of predictions?''. 

\citet{broderick2020automatic} assess the robustness of economic analyses when a fraction of data is removed. They therefore focus on the magnitude/significance of parameter estimates. 
This framing differs from our ML-centric motivation, which aims to  recover specific small subsets of data that, if removed, would change  a particular prediction (and so might support contestability). 

\vspace{0.5em}
\noindent {\bf Robustness of data analyses to dropping training data} 
In the process of review it was brought to our attention that \citet{broderick2020automatic} addressed a closely related problem to what we have considered here, albeit from a quite different motivating perspective---namely assessing the sensitivity of econometric analyses to removals of small subsets of data.
It turns out that the algorithm that was (independently) proposed by \citet{broderick2020automatic} in that work is similar to Algorithm \ref{alg:alg1}.
The present effort is novel in our focus on machine learning, and specifically on identifying minimal subsets of training data which would flip a particular prediction if removed prior to training. 

\vspace{0.5em}
\noindent{\bf Minimal feature set removal} Another related line of work concerns a natural complement to the problem we have considered: Instead of identifying a minimal set of \emph{instances} to remove in order to change a prediction, the idea is to find a minimal subset of \emph{features} such that, if these were set to uninformative values, a particular prediction would change \cite{harzli2022minimal}.
Work on \emph{counterfactual examples} has similarly sought to identify minimal (feature) edits to instances that would change the associated label \cite{kaushik2019learning}.


\vspace{0.5em}
\noindent{\bf Datamodeling} Recent work on \emph{datamodeling} \cite{pmlr-v162-ilyas22a} provided a generalized framework for analyzing model behavior as a function training data. 
This approach entails \emph{learning to estimate} (via a parameterized model) changes we would anticipate observing for a particular instance if the model had been trained on some subset of the original training set. 
This approach is flexible, and one thing it permits is identifying the \emph{data support} of a particular prediction for $x_t$, i.e., what we have called $\mathcal{S}_t$ (4.1.1 in \citealt{pmlr-v162-ilyas22a}). 
Furthermore, this method is not restricted to the simple regularized linear models we have considered here.
However, this comes with the downside of high computation costs: One needs to re-train the algorithm being modeled many times with different training data subsets to yield a ``training set'' to be used to estimate model behavior under conterfactual training sets. 
The main comparative advantage of our more focused approach is therefore relative computational efficiency.

\vspace{0.5em}
\noindent{\bf Contestability} \citep{vaccaro2019contestability,almada2019human} in ML is the idea that individuals affected by a prediction ought to be able to challenge this determination, which may require parties to ``marshal evidence and create counter narratives that argue precisely why they disagree with a conclusion drawn by an AI system'' \cite{hirsch2017designing}.
The right to contestability is in some cases enshrined into law \citep{almada2019human}.
Identifying $\mathcal{S}_t$ for review by an individual affected by the  prediction $\hat{y}_t$ may constitute a concrete mechanism for contestation. 

\section{Conclusions}

In the context of binary text classification, we investigated the problem of identifying a minimal set of training points $\mathcal{S}_t$ such that, if excluded from training, the prediction for test instance $x_t$ would flip. 
We proposed two relatively efficient algorithms for this---both using approximate group influence \cite{IF,koh2019accuracy}---and showed that for regularized linear models they can often find relatively small $\mathcal{S}_t$.
We provided empirical evidence that this captures uncertainty in a way that is somewhat complementary to predicted probabilities, and may serve as a mechanism to support \emph{contestability}, by allowing individuals to review (and dispute) instances in $\mathcal{S}_t$.

\section*{Limitations} 

A key limitation of this work is that we have restricted analysis to regularized linear models with convex loss. 
We leave extension and evaluation of the proposed methods for more complex models to future work.
Indeed, our hope is that this initial effort inspires further work on the problem of identifying minimal train sets which would overturn a specific prediction if removed.

More conceptually, the implications of finding a small subset $\mathcal{S}_t$ are not entirely clear. 
Intuitively, small sets would seem to indicate fragility, but we have not formalized or evaluated this further.  
Moreover, there may in certain cases exist \emph{multiple} (distinct) subsets $\mathcal{S}_t$, such that removing any of these subsets would flip the prediction for $x_t$. 
This would complicate the process of contestation envisioned. 
Furthermore, assuming a stochastic parameter estimation method (e.g., SGD) the composition of $\mathcal{S}_t$ may depend on the arbitrary random seed, similarly complicating the interpretation of such sets.

\section*{Acknowledgements}

We thank our anonymous EACL reviewers for helpful feedback, especially with respect to relevant related work. 
We also thank Gautam Kamath for highlighting connections to the \emph{datamodeling} work \cite{pmlr-v162-ilyas22a}. 
This work was supported in part by the Army Research Office (W911NF1810328), and in part by the Overseas Research Fellowship under the University of Hong Kong.

\section*{Ethics Statement}

Models are increasingly used to make (or aid) decisions that directly affect individuals.
In addition to the broader (potential) ``interpretability'' afforded by recovering small sets of training data that would change a prediction if removed, this may provide a new mechanism for individuals to contest such automated decisions, specifically by disputing this set of training data in some way. 
However, our proposed method only finds  training points that highly impact the model prediction for a given example; these may or may not be noisy or problematic instances. 
Human judgement is required to assess the accuracy and relevancy of the instances in $\mathcal{S}_t$.


A broader view might be that classification models are simply not appropriate for the kinds of sensitive applications we have used as motivation here.
The use of (semi-)automated methods for essay grading, e.g., has long been debated \cite{hearst2000debate}. 
One might argue that rather than trying to provide mechanisms to contest ML predictions, a better choice may be not to use models in cases where these would be necessary at all. 
We are sympathetic to this view, but view the ``appropriateness'' of ML for a given problem as a spectrum; contestability may be useful even in ``lower stakes'' cases.
Moreover, the general problem we have introduced of identifying small training sets which can by themselves swing predictions, and the corresponding methods we have proposed for recovering these, may be of intrinsic interest beyond contestability (e.g., as an additional sort of model uncertainty).



\bibliography{custom}
\bibliographystyle{acl_natbib}
\clearpage

\appendix

\setcounter{table}{0}
\setcounter{figure}{0}
\renewcommand{\thetable}{A\arabic{table}}
\renewcommand\thefigure{A.\arabic{figure}}

\section{Appendix}
\label{sec:appendix}

\subsection{Dataset Statistics and Predictive Performance}
\label{section:data-stats}

\begin{table}
\small
\centering 
\begin{tabular}{l l l l}
\hline
{\textbf{Dataset}} & {\textbf{\texttt{\#} Train}} & {\textbf{ \texttt{\#} Test}} & {\textbf{\% Pos}} \\
\hline 
Movie reviews          & 6920 & 872 & 0.52  \\
Essays                 & 11678 & 1298 & 0.10    \\
Emotion                & 9025  & 1003 & 0.53 \\
Hate speech            & 9632 & 1071 & 0.11   \\
Tweet sentiment        & 18000 & 1000 & 0.50  \\
\hline
\end{tabular}
\caption{Text classification dataset statistics.}
\label{table:dataset-info}
\end{table}

We present basic statistics describing our text classification datasets in Table \ref{table:dataset-info}. For the tweet sentiment dataset, we randomly sampled 19000 points from the 1600000 points to make experiments feasible. 
For reference, we also report the predictive performance realized by the models considered on the test sets of these corpora in Table \ref{table:model-performance}. 

\begin{table}
\small
\centering 
\begin{tabular}{l l l l}
\hline
{\textbf{Models}} & {\textbf{Accuracy}} & {\textbf{F1-score}} & {\textbf{AUC}} \\
\hline 
 \multicolumn{3}{c}{\emph{Movie reviews}} \\
BoW & 0.79 & 0.80 & 0.88\\
BERT & 0.82 & 0.83 & 0.91\\
\multicolumn{3}{c}{\emph{Essays}} \\
BoW      & 0.97 & 0.80 & 0.99\\
BERT &  0.97 & 0.84 & 0.99\\
 \multicolumn{3}{c}{\emph{Emotion classification}} \\
BoW  & 0.77 & 0.79 & 0.86 \\
BERT & 0.80 & 0.82 & 0.88\\
 \multicolumn{3}{c}{\emph{Hate speech}} \\
BoW      & 0.87 & 0.40 & 0.81  \\
BERT &  0.89 & 0.63 & 0.88\\
\multicolumn{3}{c}{\emph{Tweet sentiment}} \\
BoW         & 0.70 & 0.70 & 0.75\\
BERT        & 0.75 & 0.76 & 0.84\\
\hline
\end{tabular}
\caption{The model performance respect to datasets included in the experiment.}
\label{table:model-performance}
\end{table}

\subsection{Full results}
\label{section:all-results}

Table \ref{table:algo1-flips} reports the percentages of instances for which Algorithm \ref{alg:alg1} identifies a subset $\mathcal{S}_t$ (center column), and for which this set actually flipped the prediction following removal (right column). 
Contrast this with Table \ref{table:k-flips}, which reports the same for the proposed iterative approach in Algorithm \ref{alg:iterative}.

\begin{figure*}
\centering
\begin{subfigure}{.35\linewidth}
    \centering
    \includegraphics[width=\textwidth]{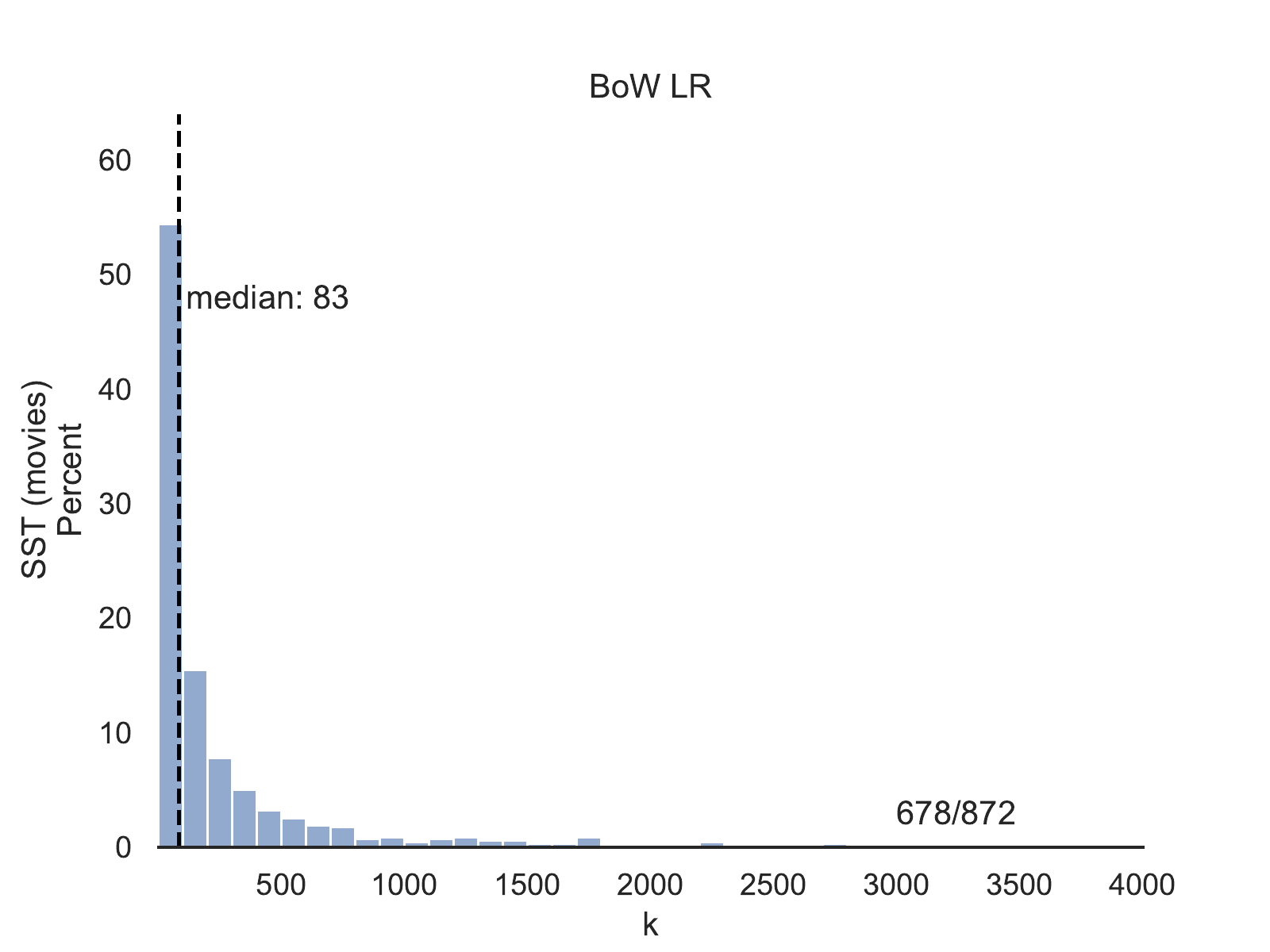}
\end{subfigure}
\hspace{1em}
\begin{subfigure}{.35\linewidth}
    \centering
    \includegraphics[width=\textwidth]{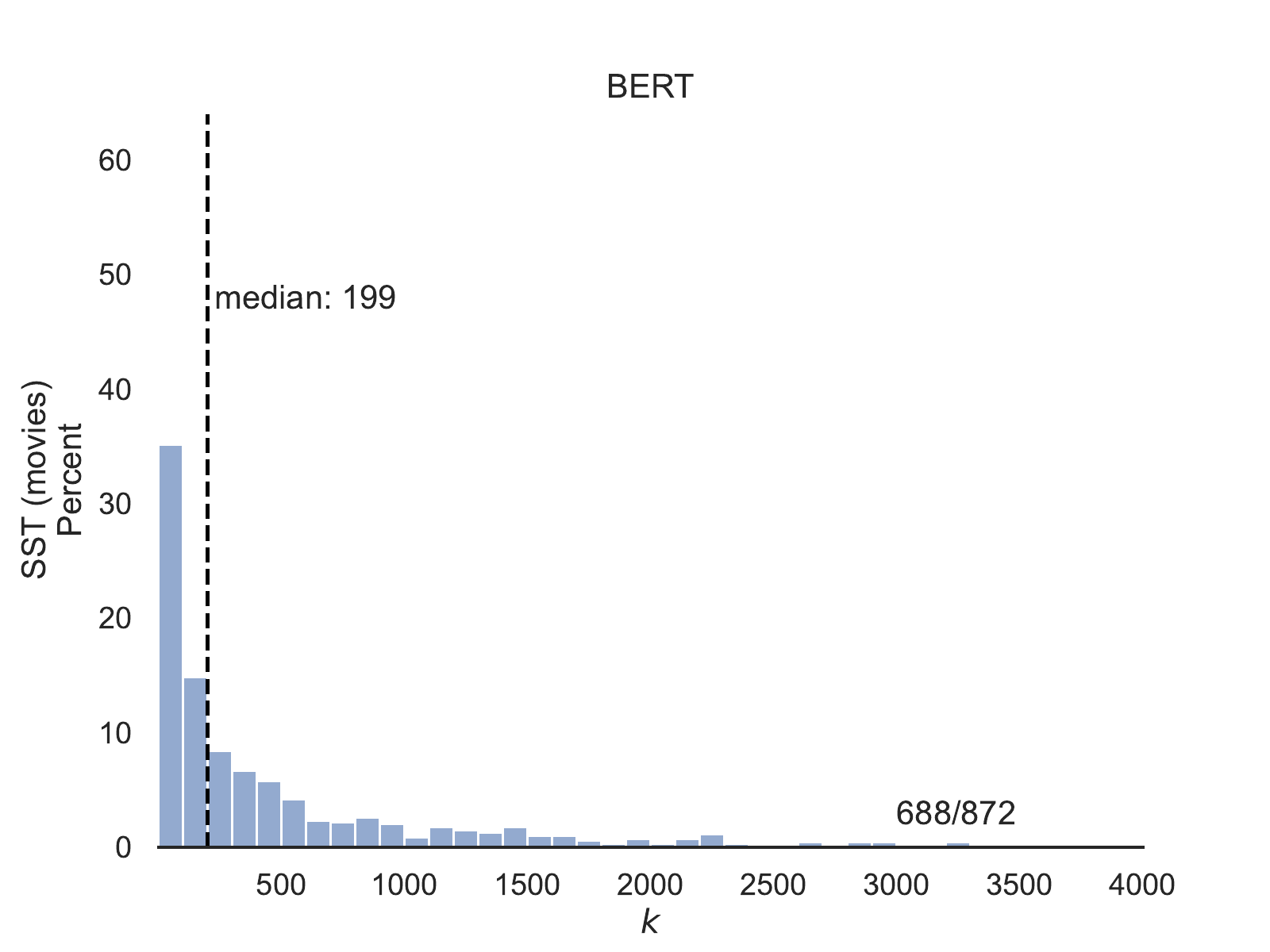}
\end{subfigure}

\hspace{1em}
\begin{subfigure}{.35\linewidth}
    \centering
    \includegraphics[width=\textwidth]{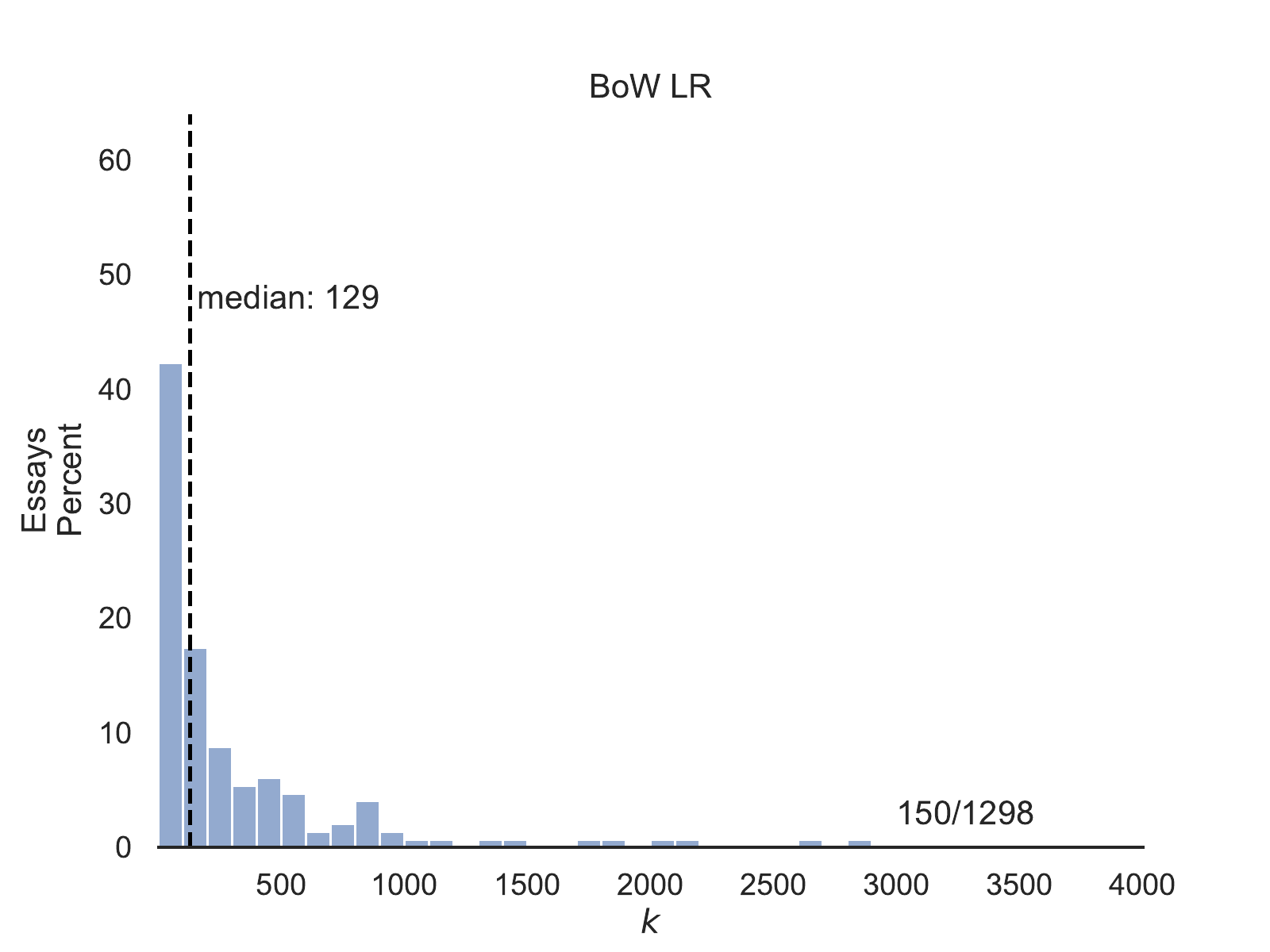}
\end{subfigure}
\hspace{1em}
\begin{subfigure}{.35\linewidth}
    \centering
    \includegraphics[width=\textwidth]{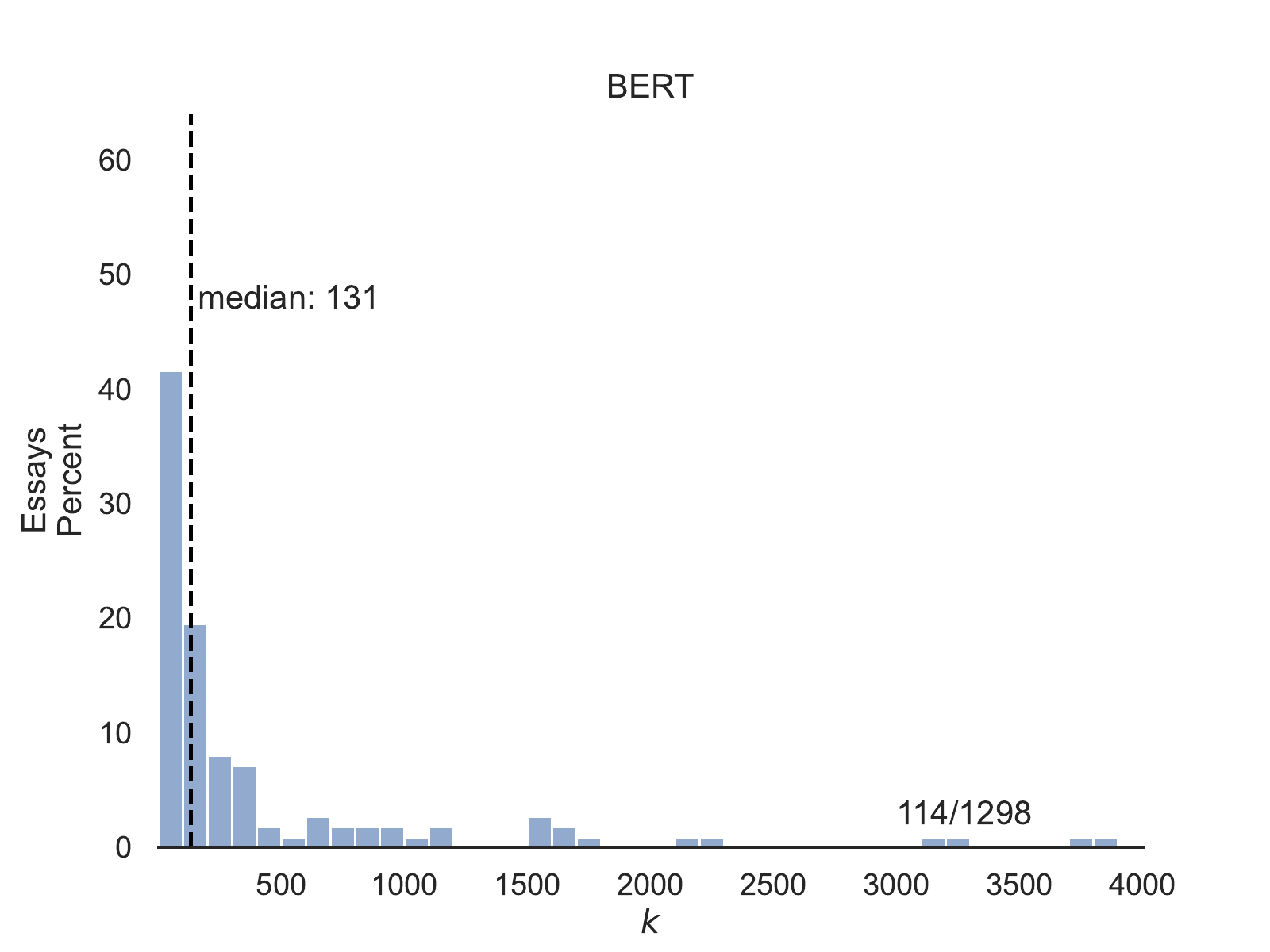}
\end{subfigure}

\hspace{1em}
\begin{subfigure}{.35\linewidth}
    \centering
    \includegraphics[width=\textwidth]{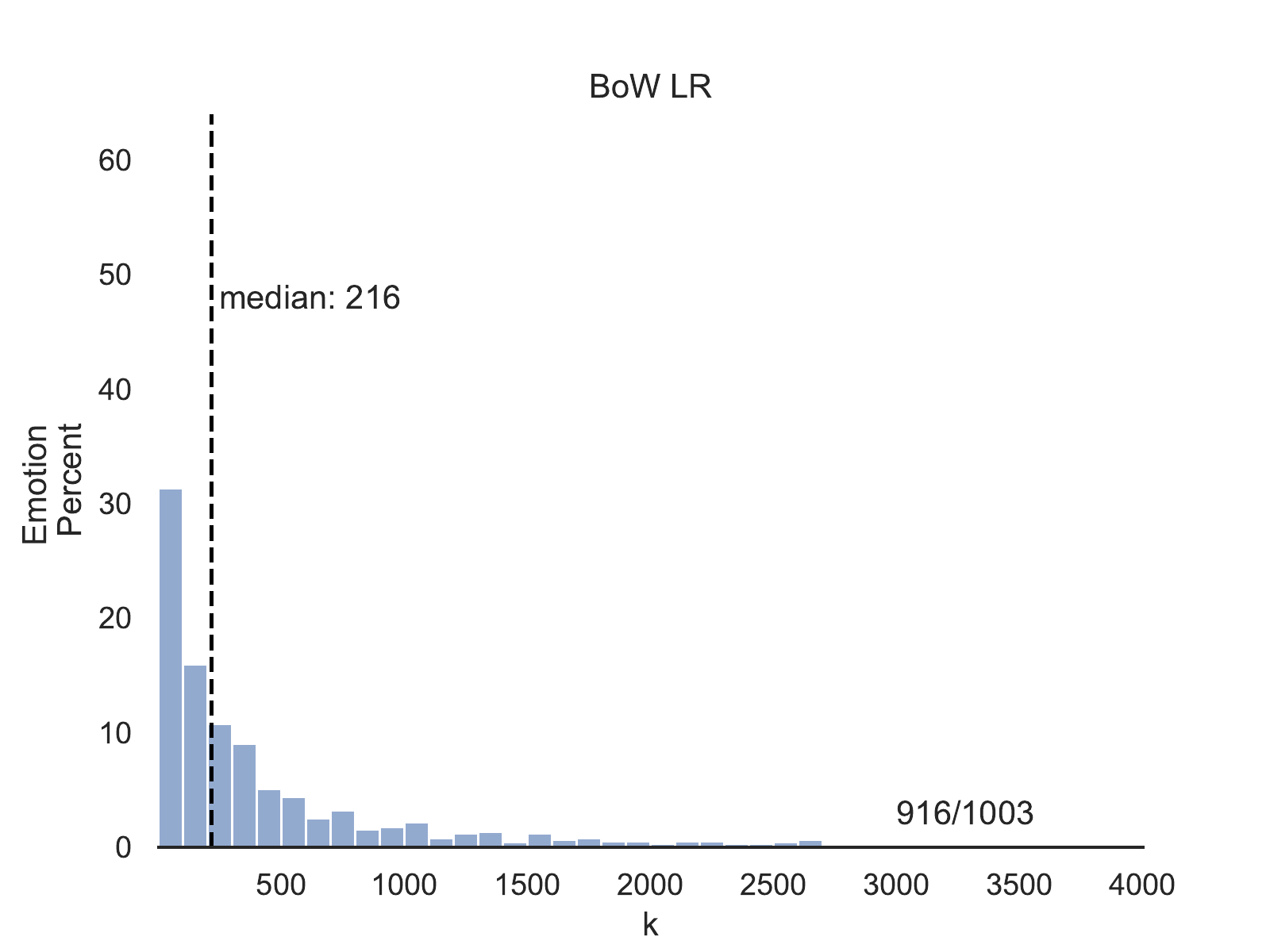}
\end{subfigure}
\hspace{1em}
\begin{subfigure}{.35\linewidth}
    \centering
    \includegraphics[width=\textwidth]{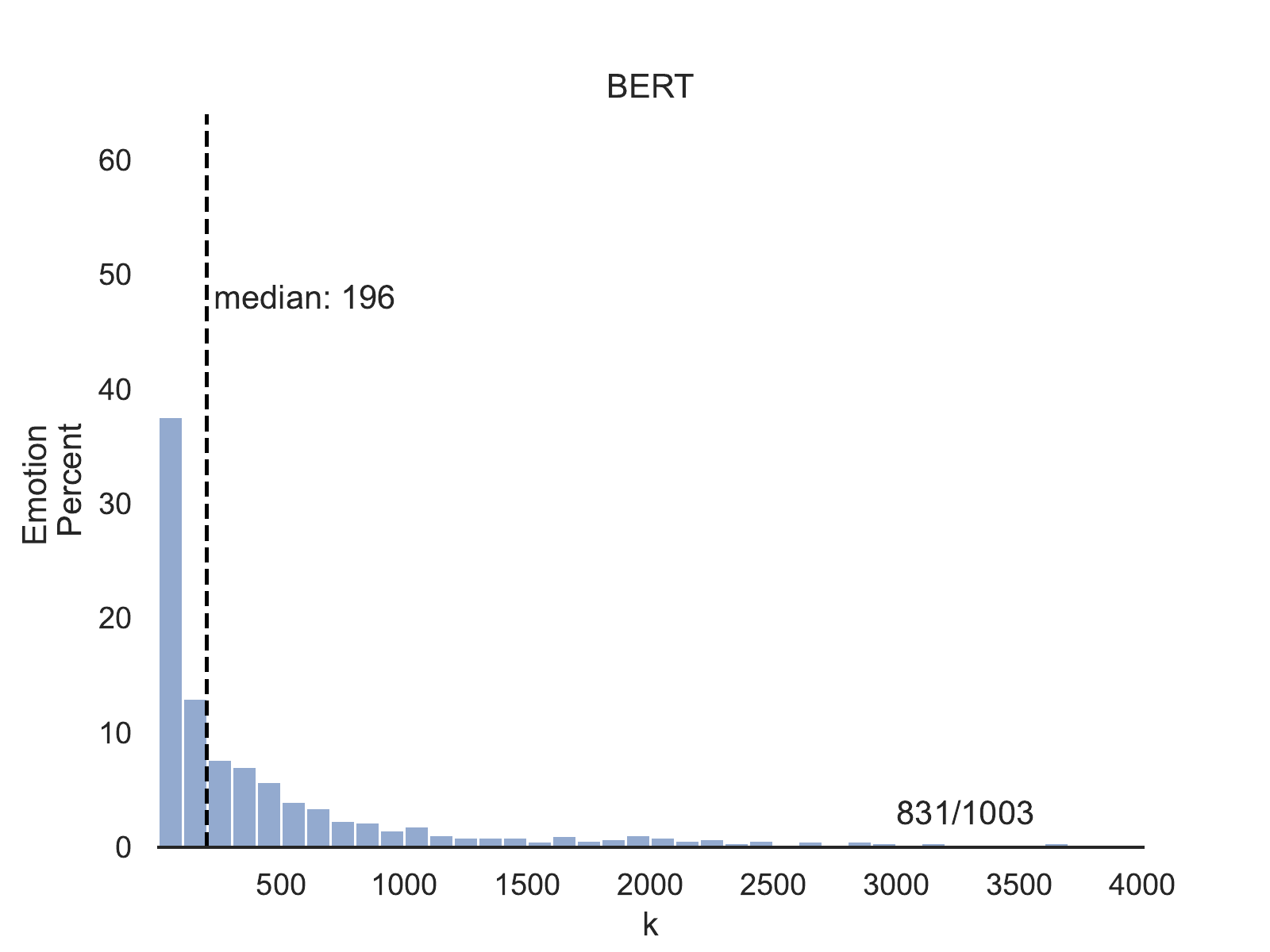}
\end{subfigure}

\hspace{1em}
\begin{subfigure}{.35\linewidth}
    \centering
    \includegraphics[width=\textwidth]{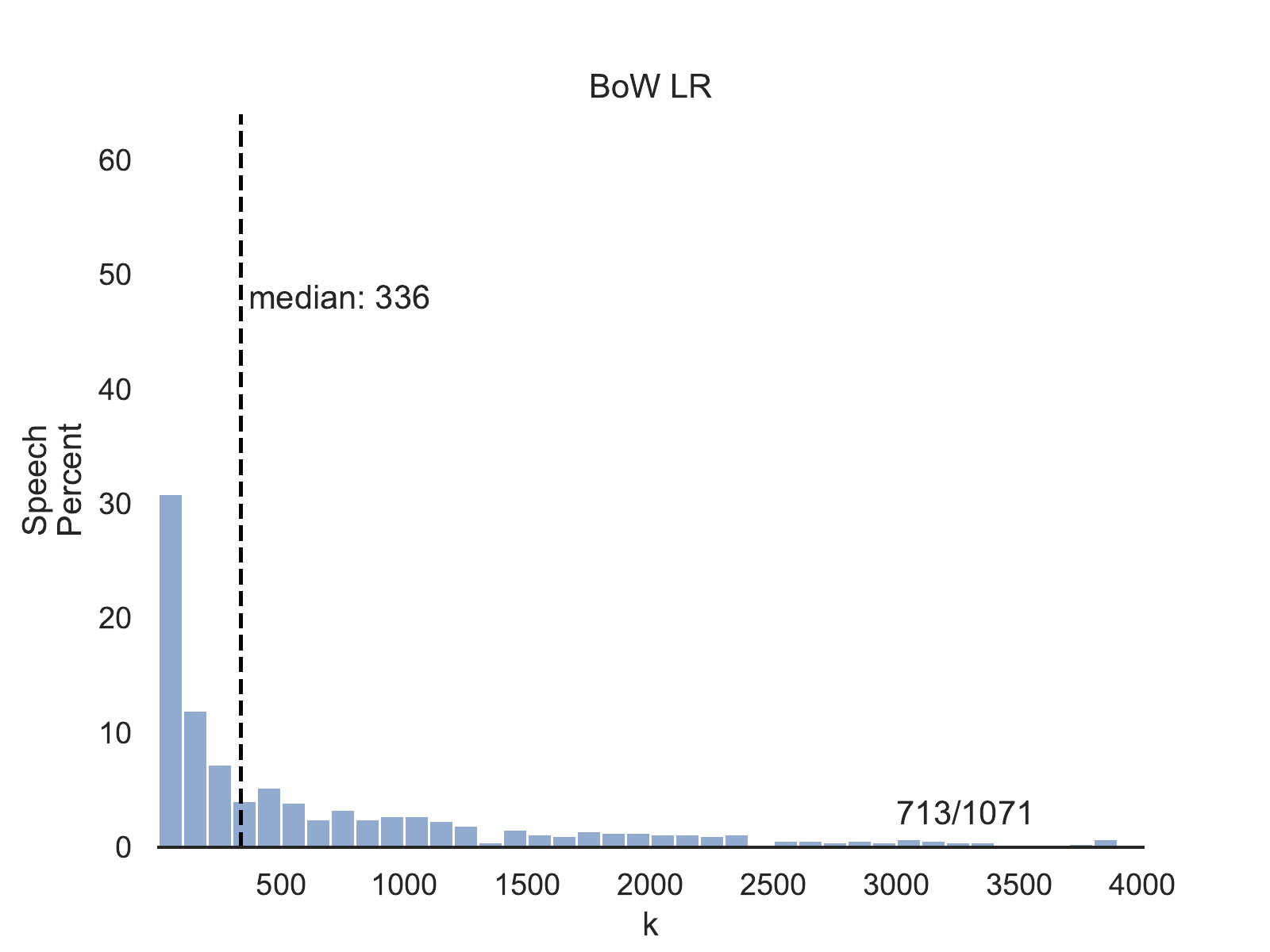}
\end{subfigure}
\hspace{1em}
\begin{subfigure}{.35\linewidth}
    \centering
    \includegraphics[width=\textwidth]{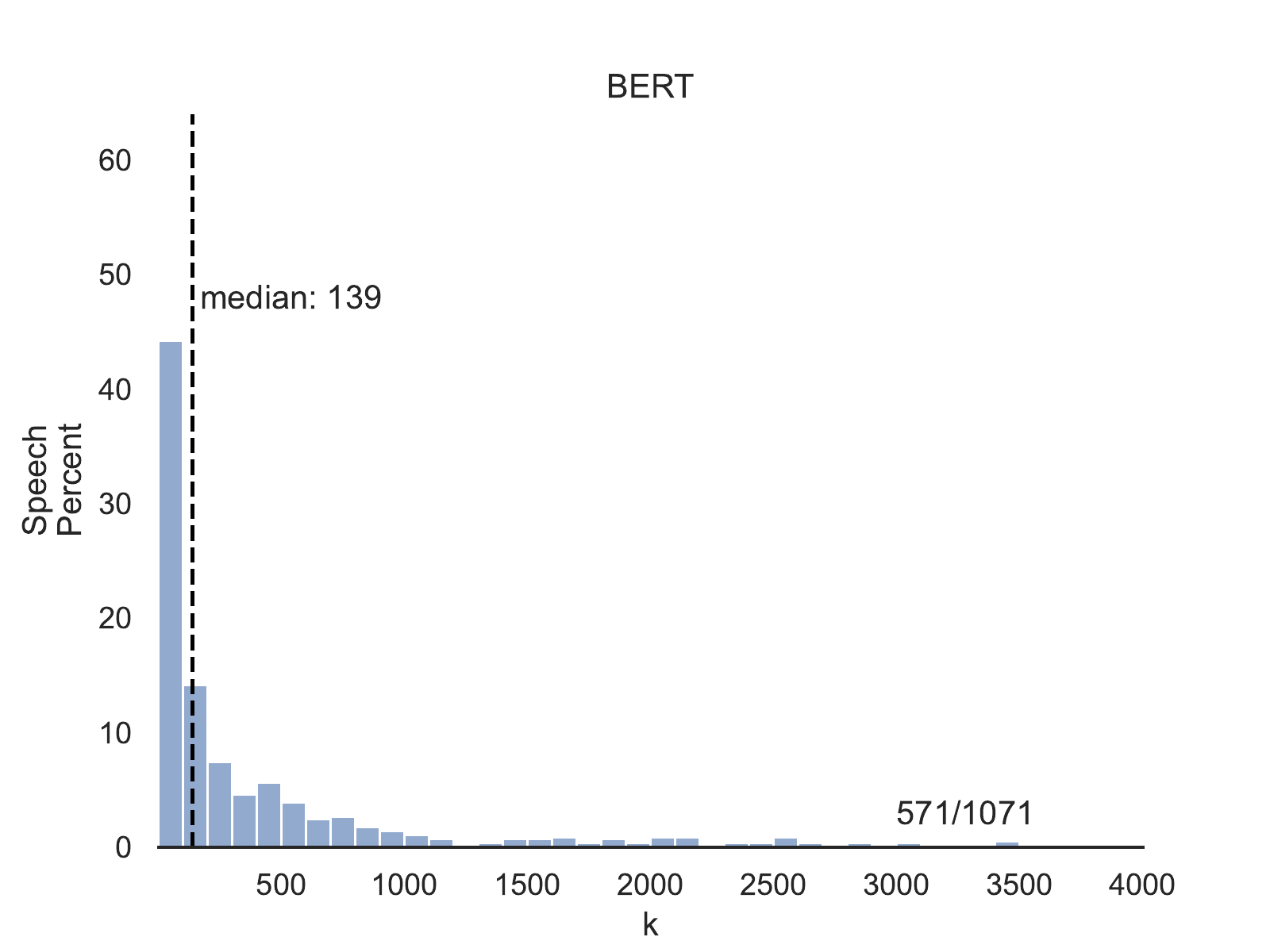}
\end{subfigure}

\hspace{1em}
\begin{subfigure}{.35\linewidth}
    \centering
    \includegraphics[width=\textwidth]{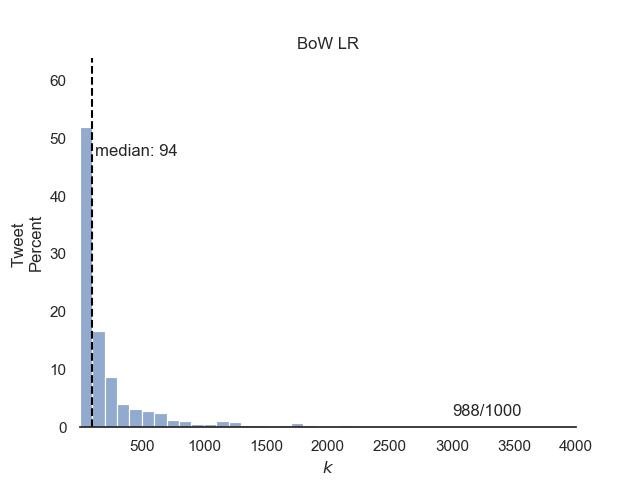}
\end{subfigure}
\hspace{1em}
\begin{subfigure}{.35\linewidth}
    \centering
    \includegraphics[width=\textwidth]{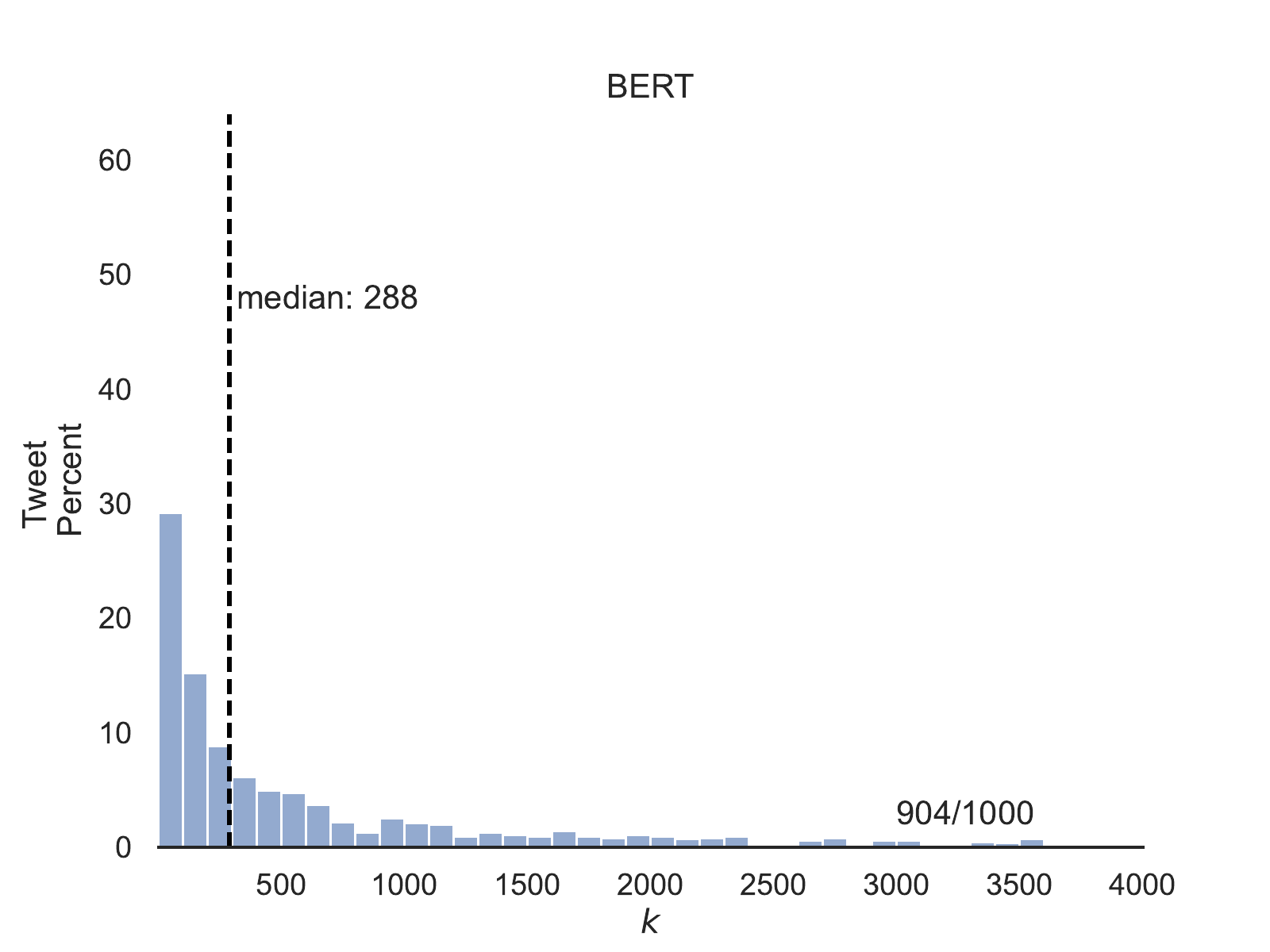}
\end{subfigure}

\caption{Histograms of $k=|\mathcal{S}_t|$ values from Algorithm \ref{alg:alg1} over the subsets of test points $x_t$ for which we were able to successfully identify a set of points $\mathcal{S}_t$ such that removing them would flip the prediction for $\hat{y}_t$.}
\label{fig:dist_alg1}
\end{figure*}

\begin{figure*}
\centering
\begin{subfigure}{.425\linewidth}
    \centering
    \includegraphics[width=\textwidth]{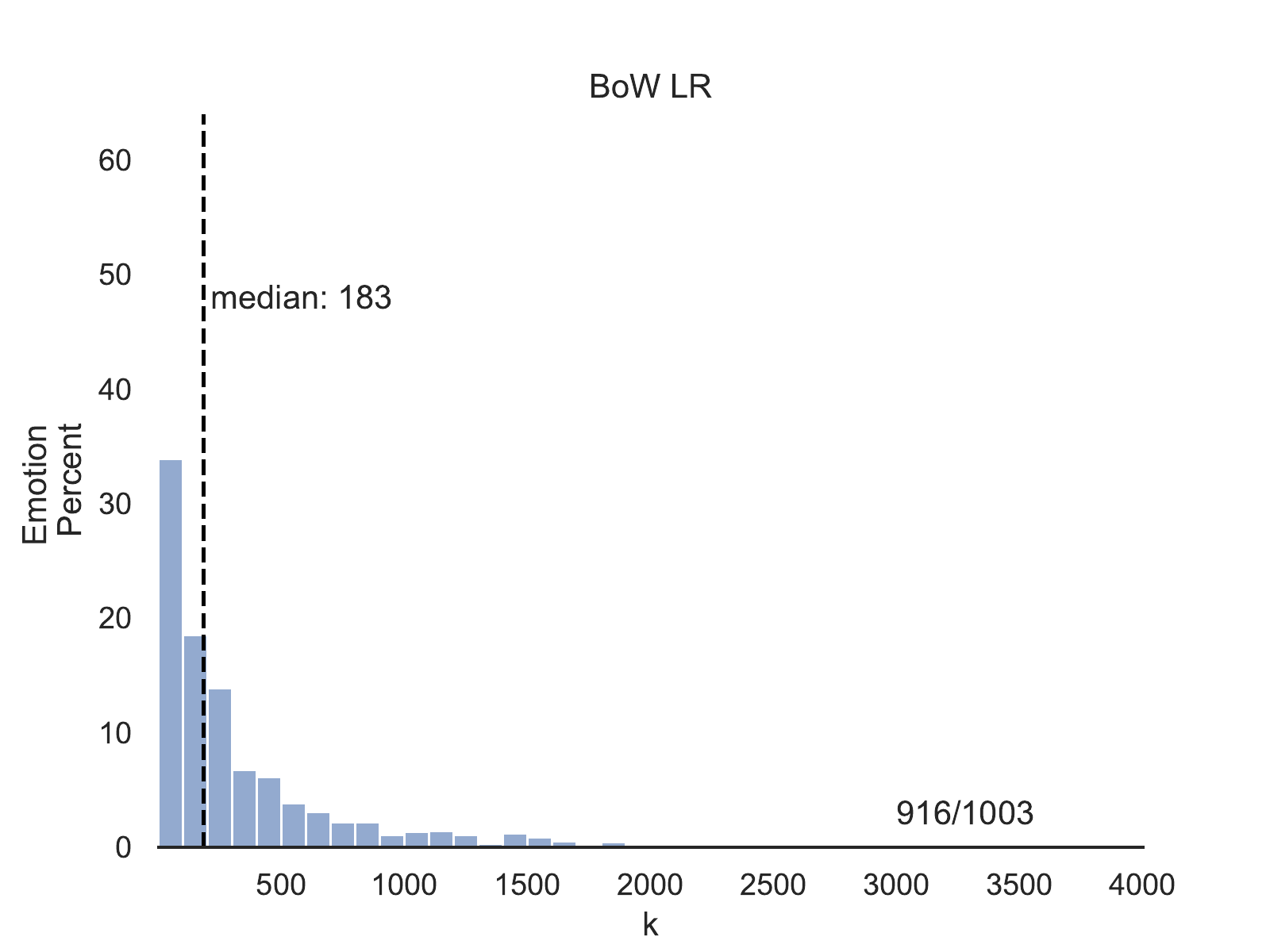}
\end{subfigure}
\hspace{1em}
\begin{subfigure}{.425\linewidth}
    \centering
    \includegraphics[width=\textwidth]{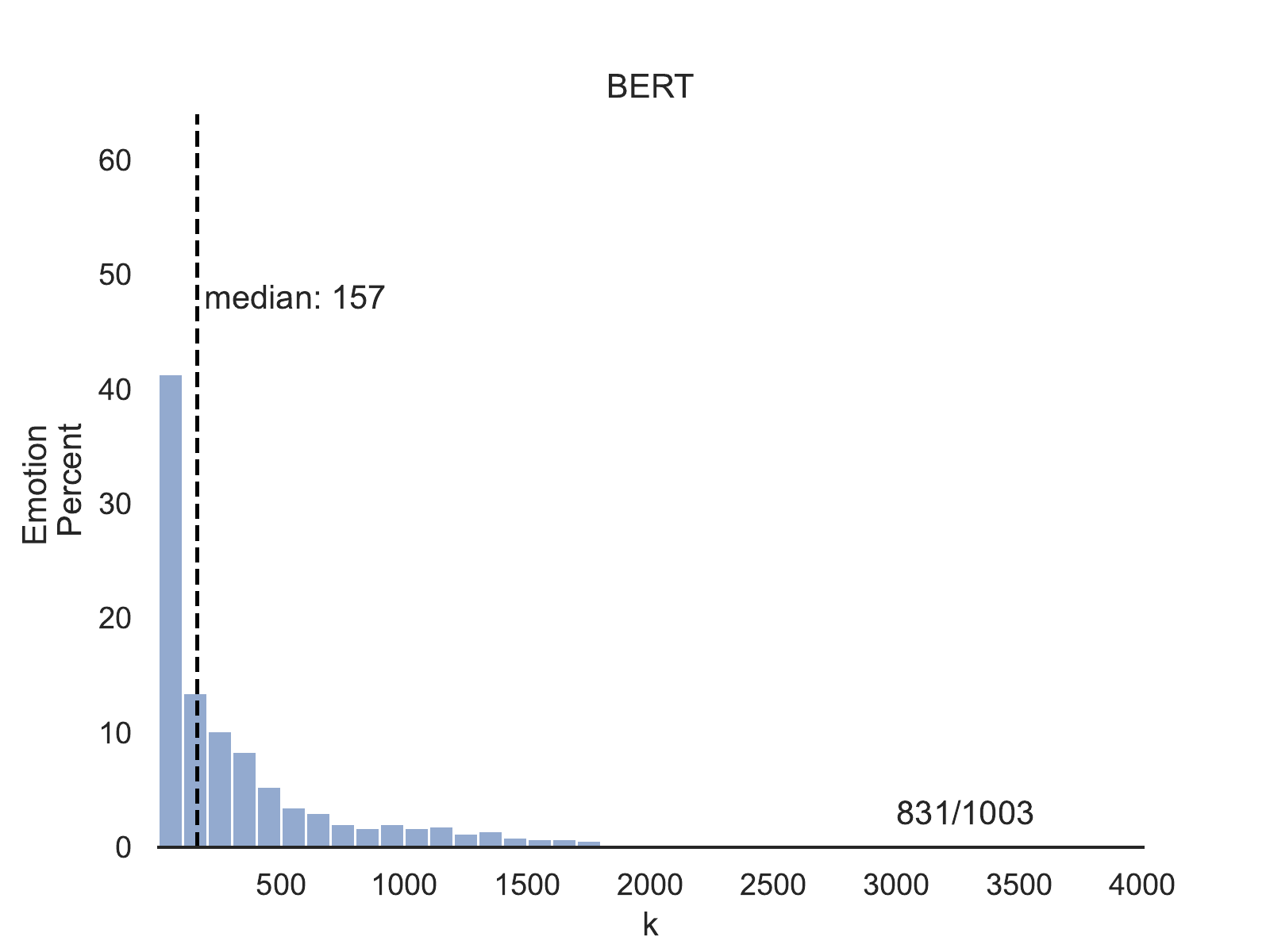}
\end{subfigure}

\hspace{1em}
\begin{subfigure}{.425\linewidth}
    \centering
    \includegraphics[width=\textwidth]{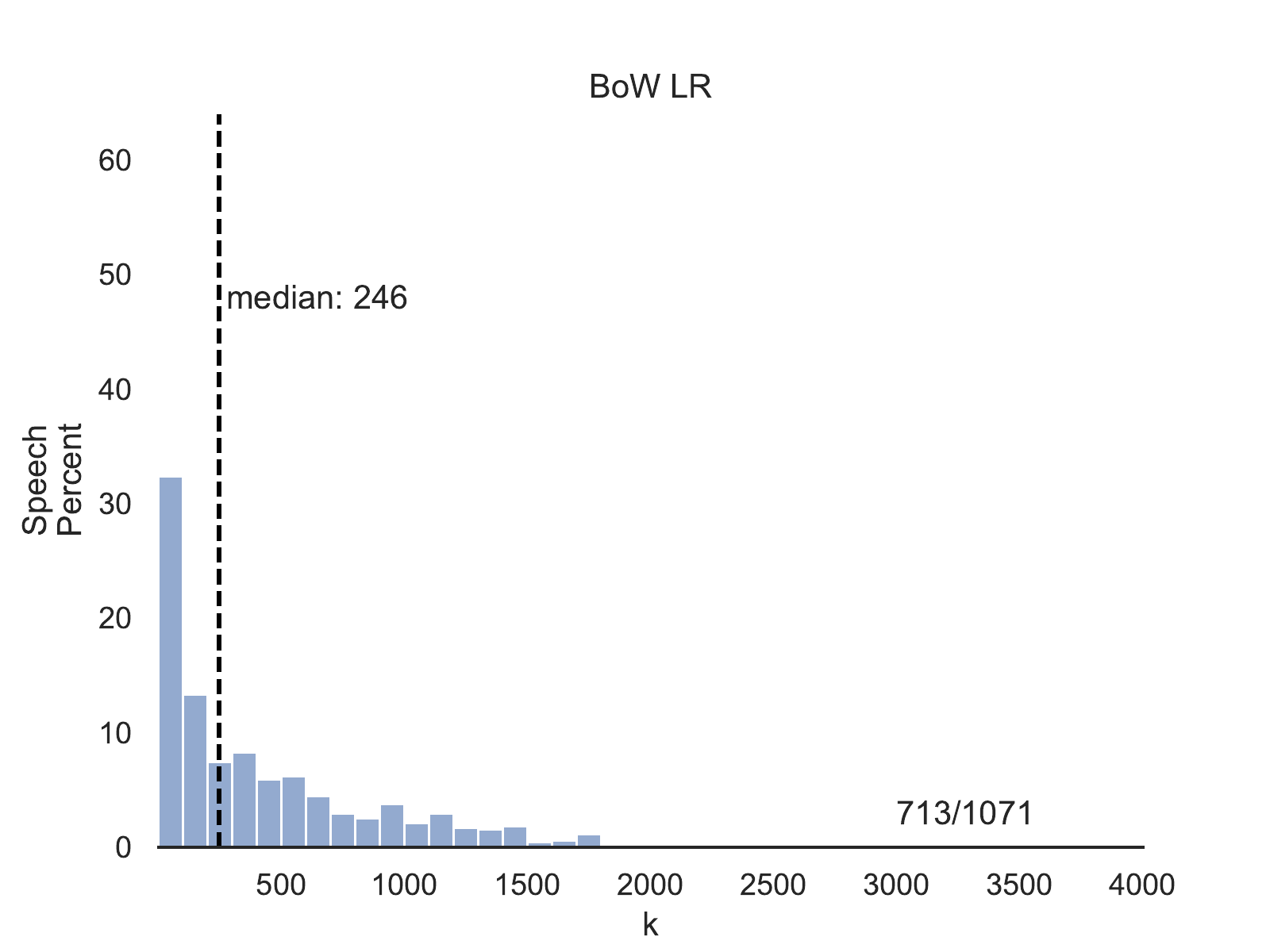}
\end{subfigure}
\hspace{1em}
\begin{subfigure}{.425\linewidth}
    \centering
    \includegraphics[width=\textwidth]{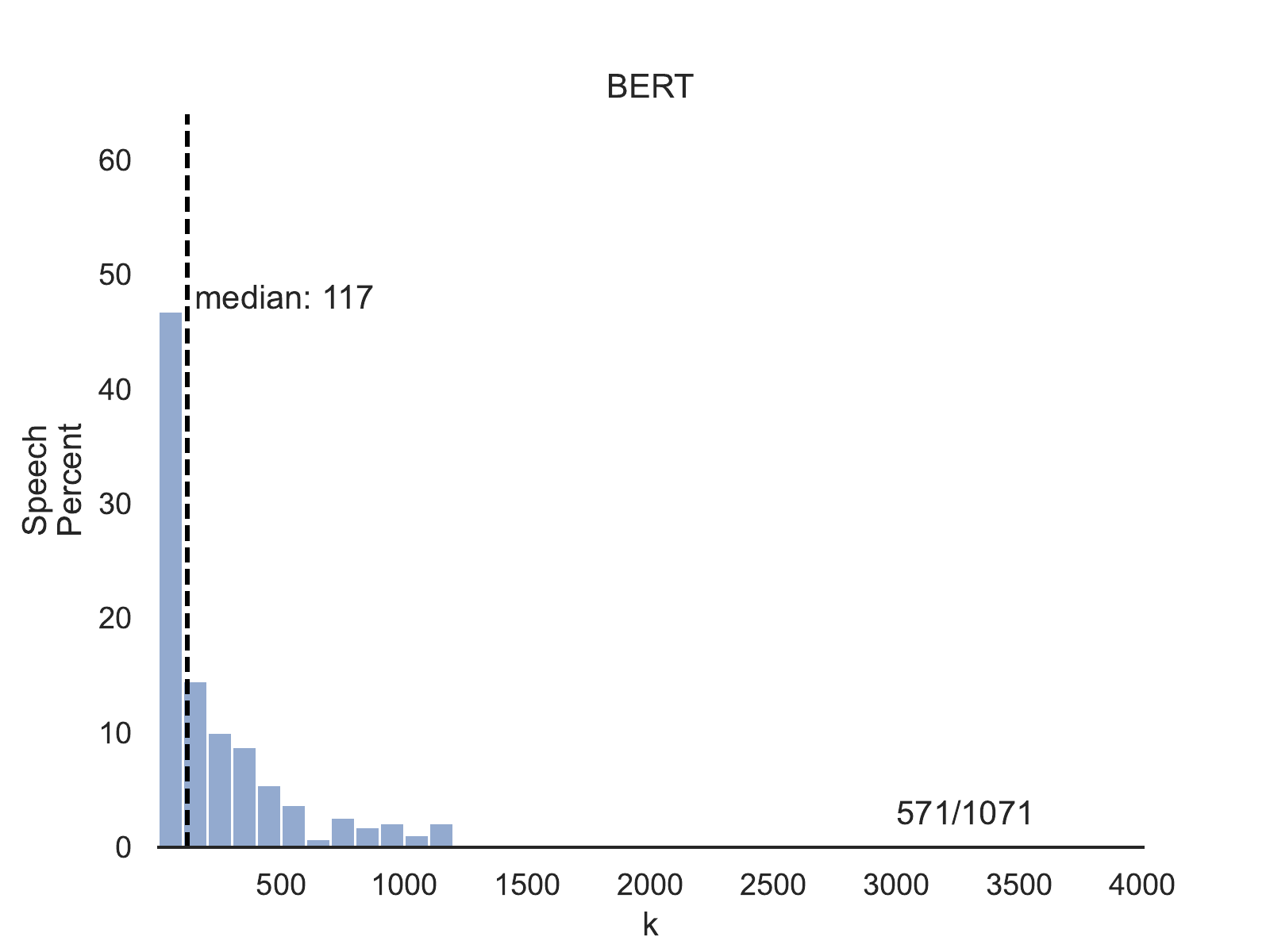}
\end{subfigure}

\hspace{1em}
\begin{subfigure}{.425\linewidth}
    \centering
    \includegraphics[width=\textwidth]{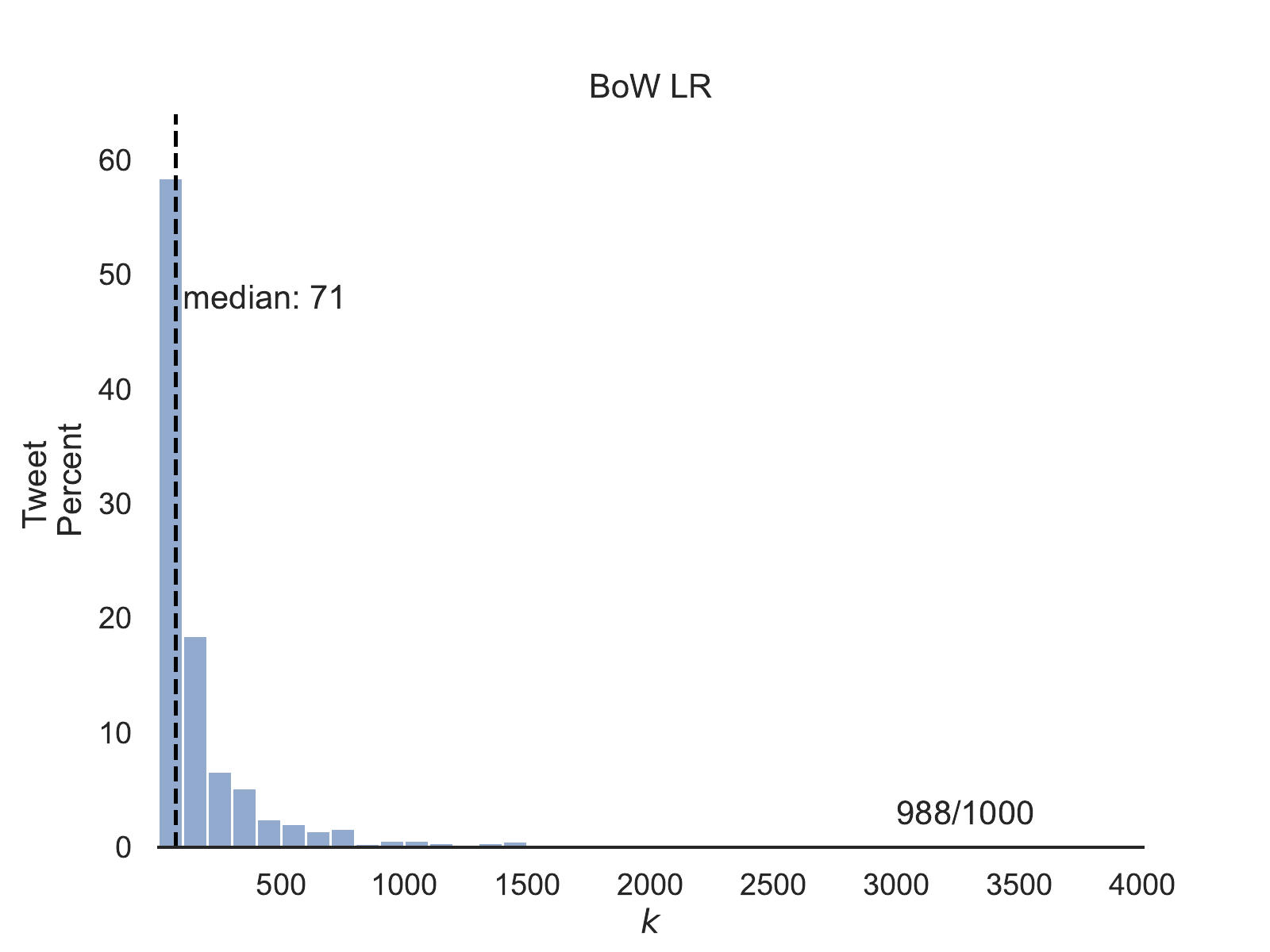}
\end{subfigure}
\hspace{1em}
\begin{subfigure}{.425\linewidth}
    \centering
    \includegraphics[width=\textwidth]{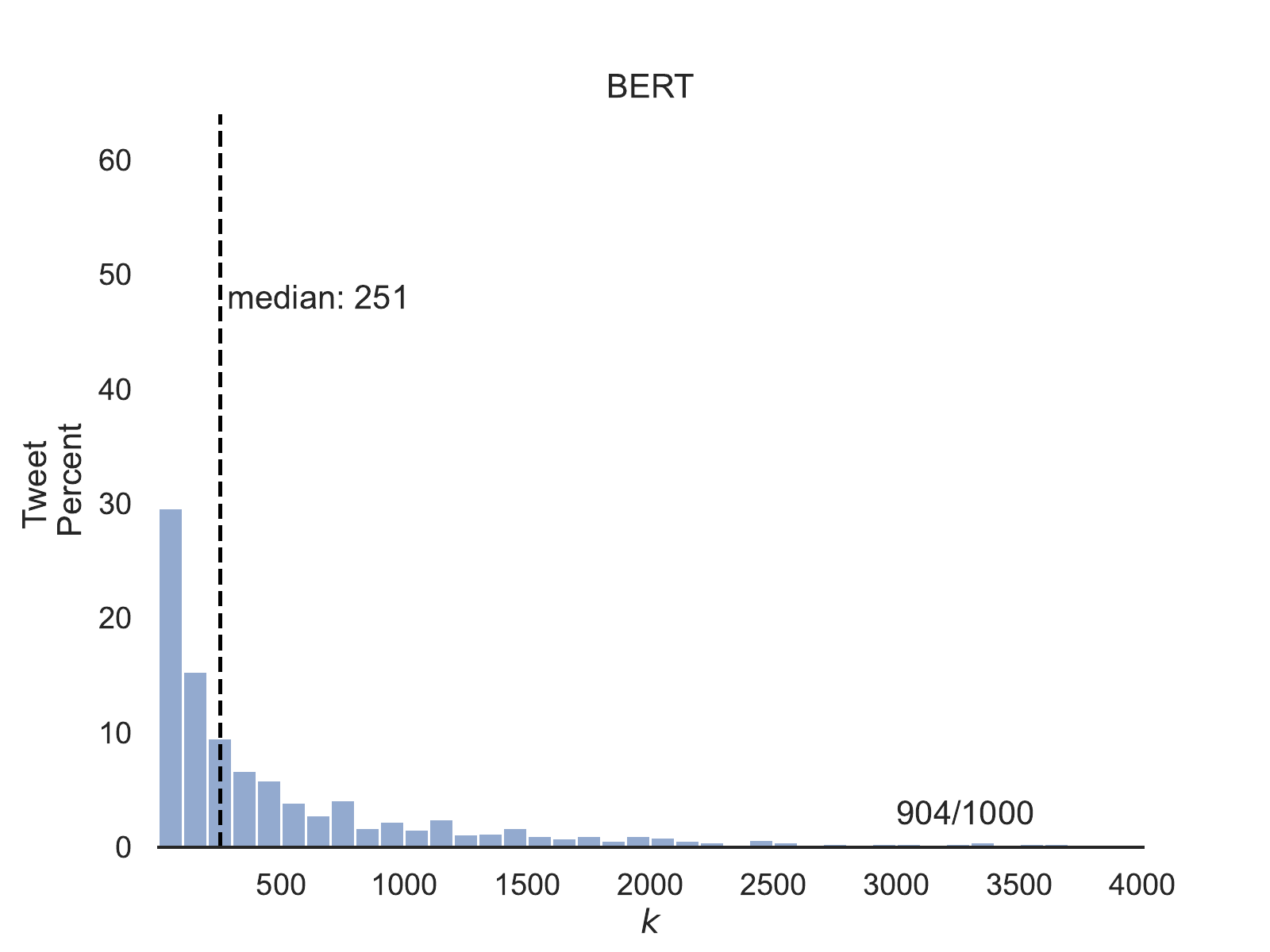}
\end{subfigure}

\caption{Histograms of $k=|\mathcal{S}_t|$ values from Algorithm \ref{alg:iterative} over the subsets of test points $x_t$ for which we were able to successfully identify a set of points $\mathcal{S}_t$ such that removing them would flip the prediction for $\hat{y}_t$.}
\label{fig:dist_alg2}
\end{figure*}

We provide histograms of $k=|\mathcal{S}_t|$ for the sets we were able to identify via Algorithm \ref{alg:alg1} in Figure \ref{fig:dist_alg1}, and the same plots for Algorithm \ref{alg:iterative} in Figure \ref{fig:dist_alg2}.

\begin{figure*}
\centering
\begin{subfigure}{.35\linewidth}
    \centering
    \includegraphics[width=\textwidth]{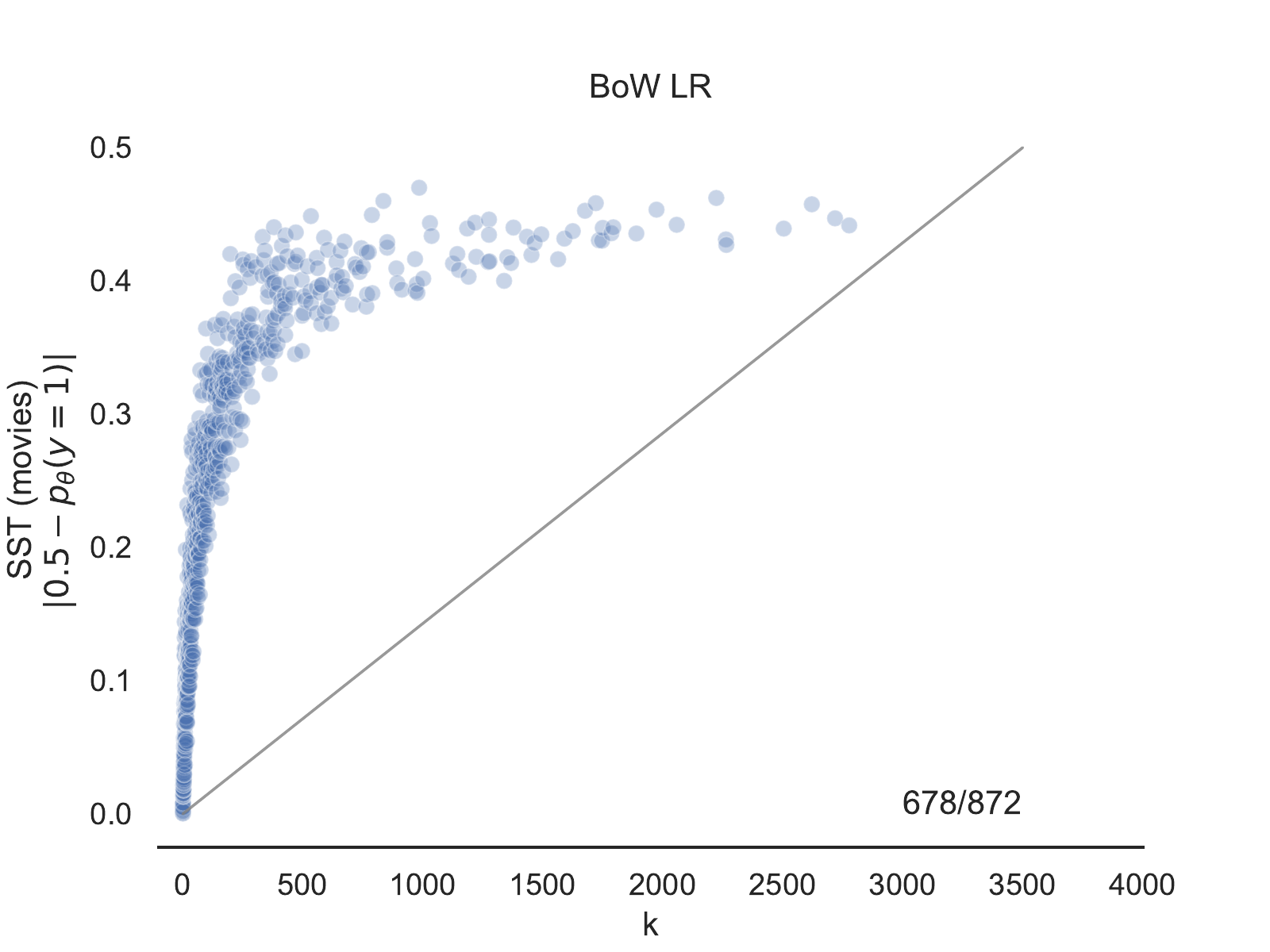}
\end{subfigure}
\hspace{1em}
\begin{subfigure}{.35\linewidth}
    \centering
    \includegraphics[width=\textwidth]{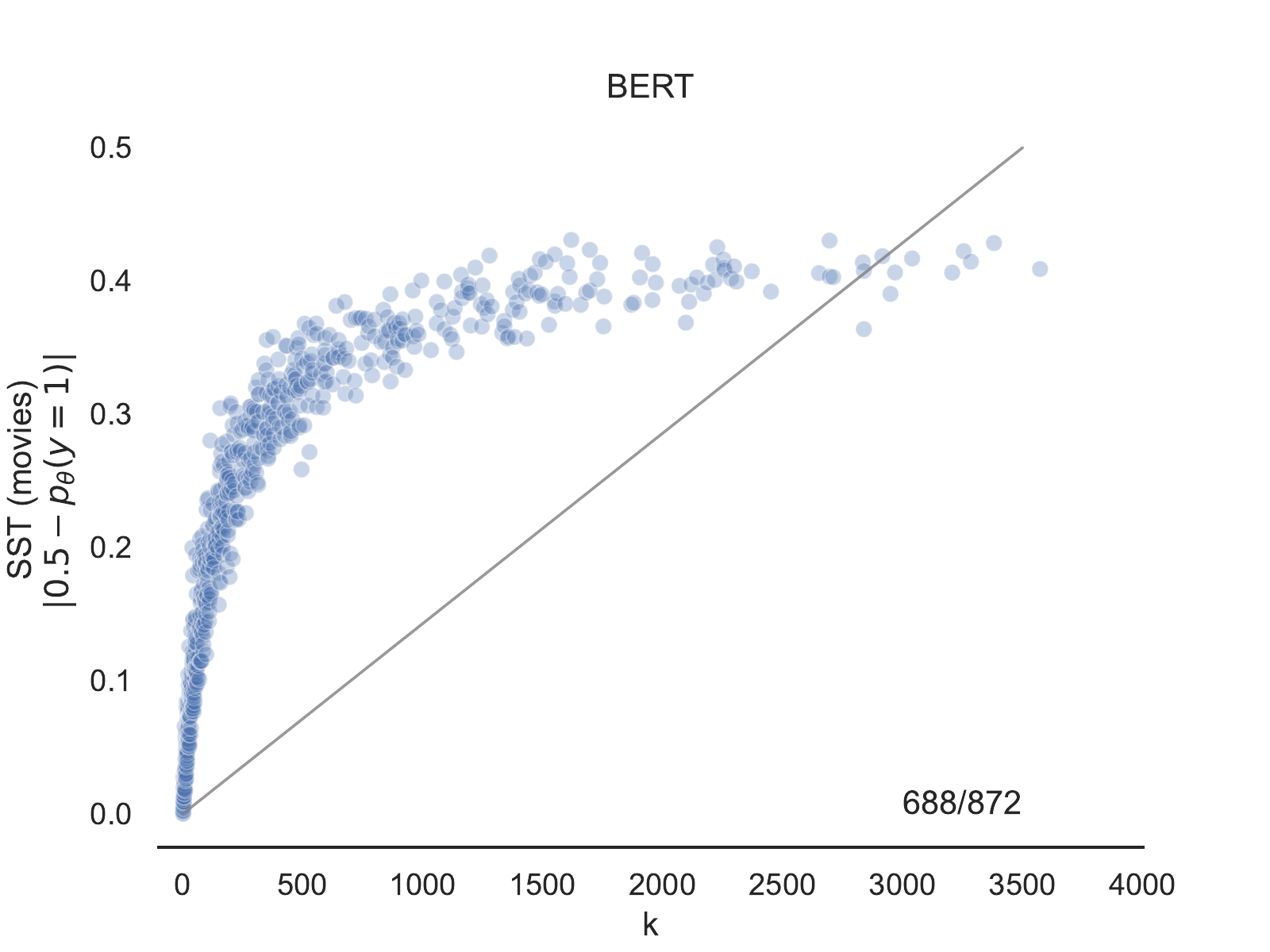}
\end{subfigure}

\hspace{1em}
\begin{subfigure}{.35\linewidth}
    \centering
    \includegraphics[width=\textwidth]{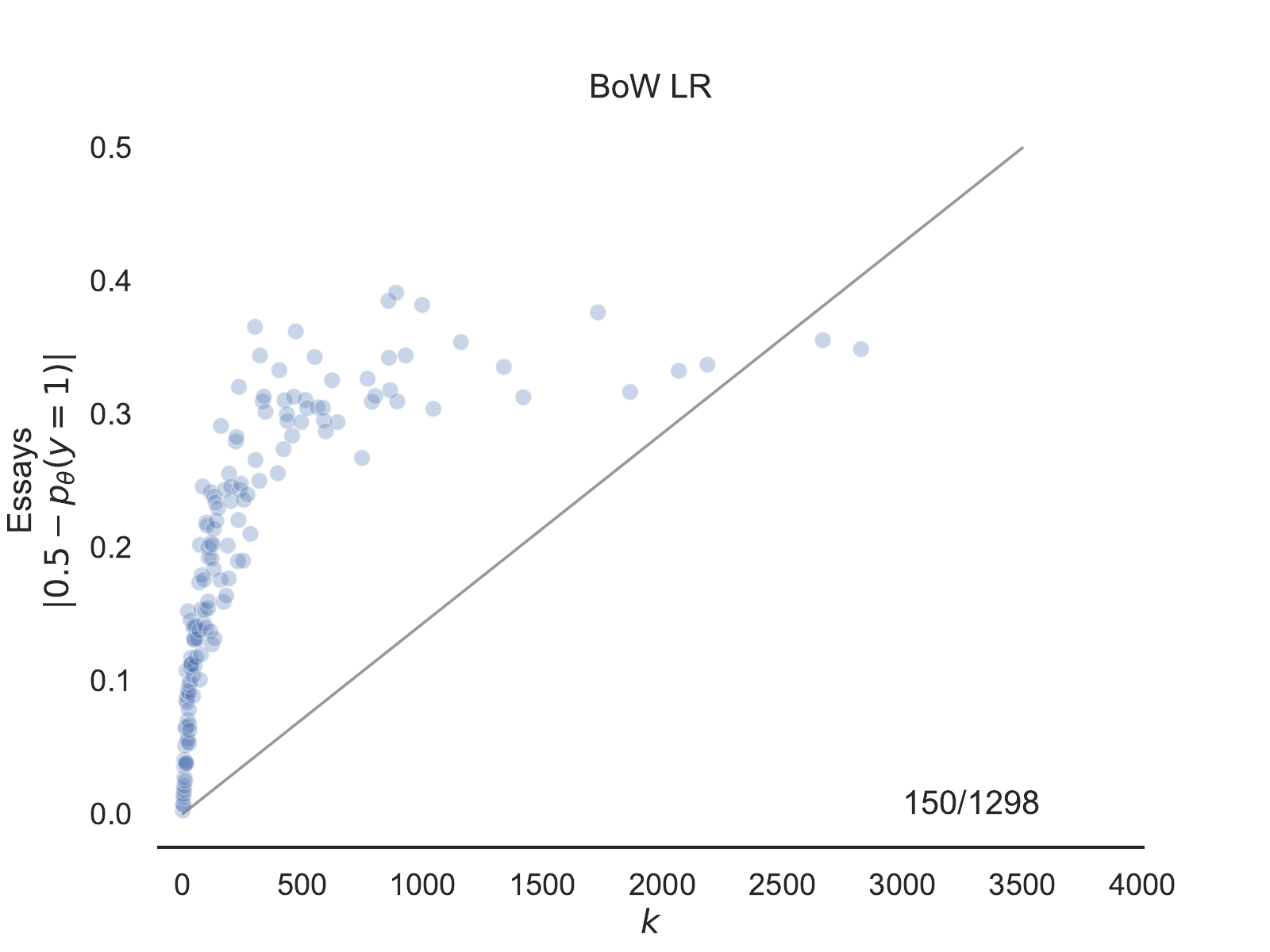}
\end{subfigure}
\hspace{1em}
\begin{subfigure}{.35\linewidth}
    \centering
    \includegraphics[width=\textwidth]{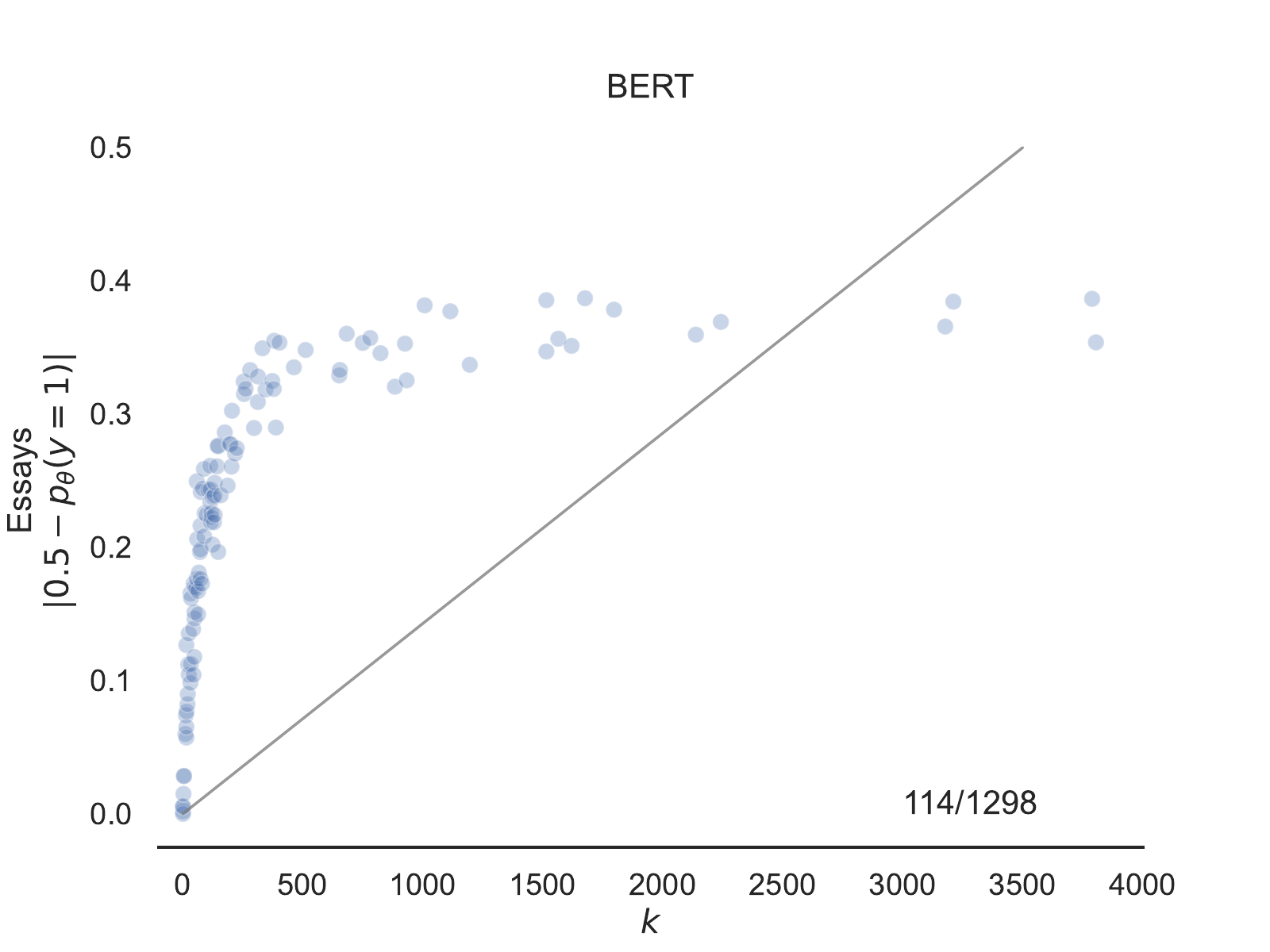}
\end{subfigure}

\hspace{1em}
\begin{subfigure}{.35\linewidth}
    \centering
    \includegraphics[width=\textwidth]{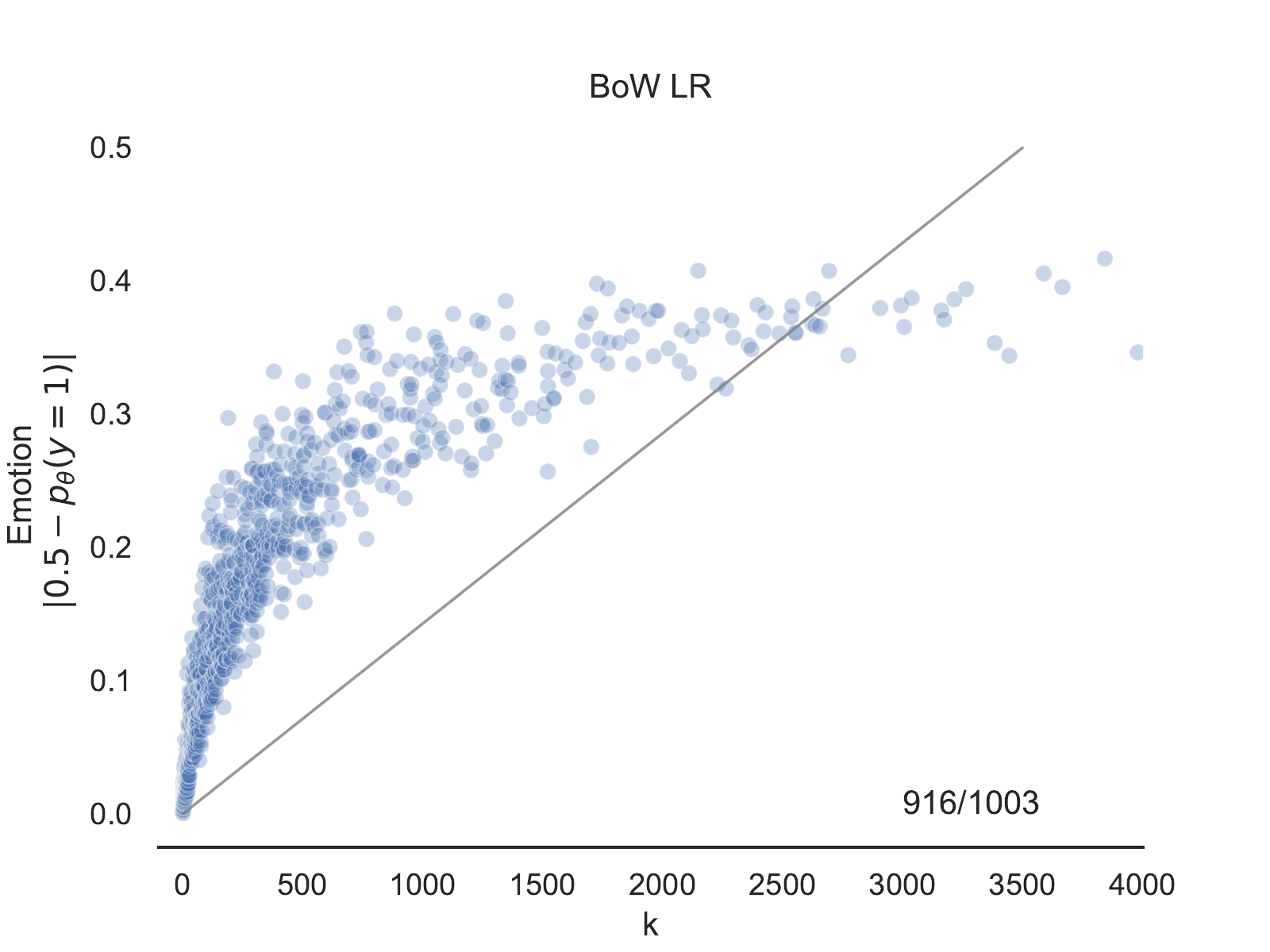}
\end{subfigure}
\hspace{1em}
\begin{subfigure}{.35\linewidth}
    \centering
    \includegraphics[width=\textwidth]{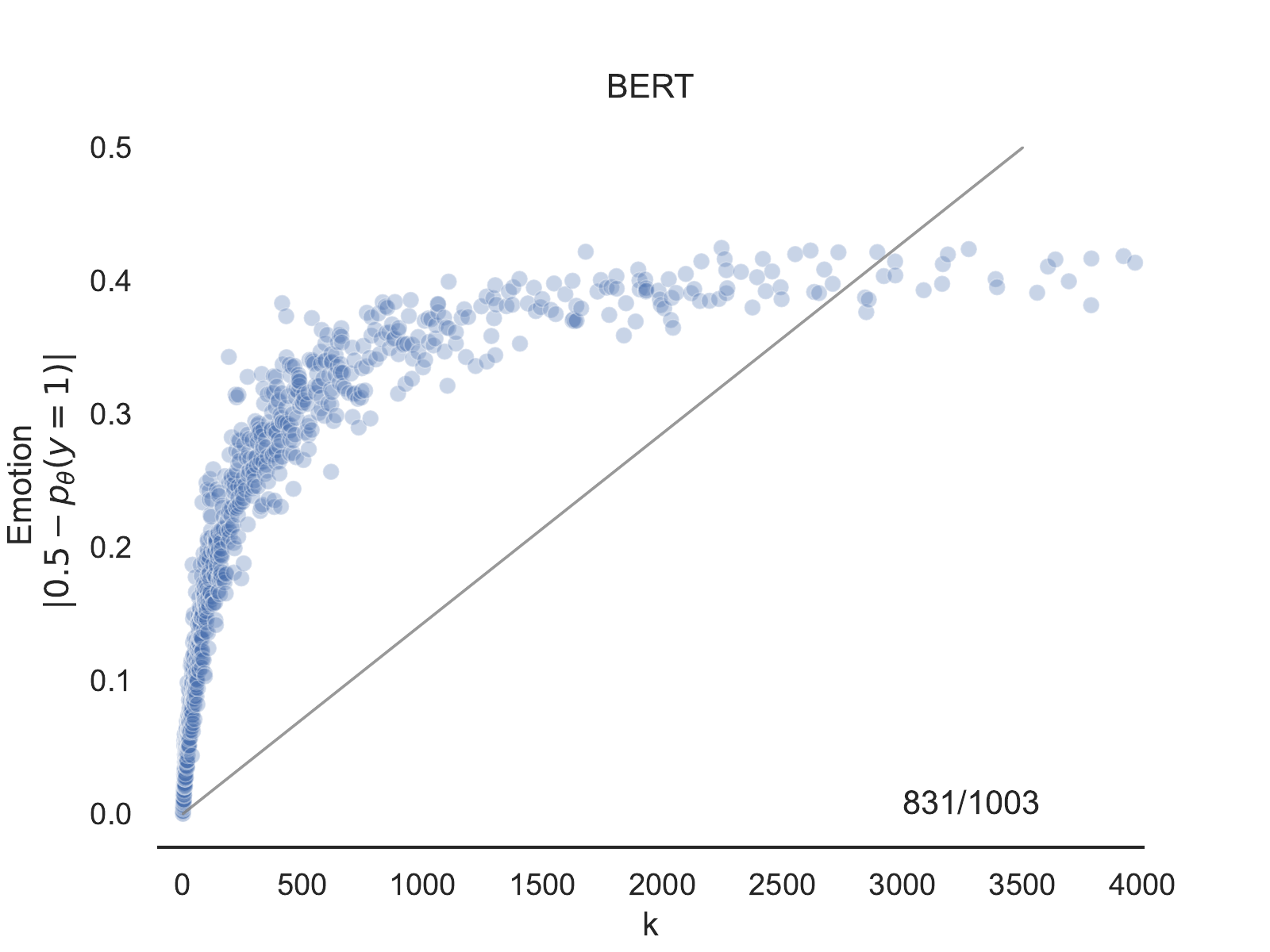}
\end{subfigure}

\hspace{1em}
\begin{subfigure}{.35\linewidth}
    \centering
    \includegraphics[width=\textwidth]{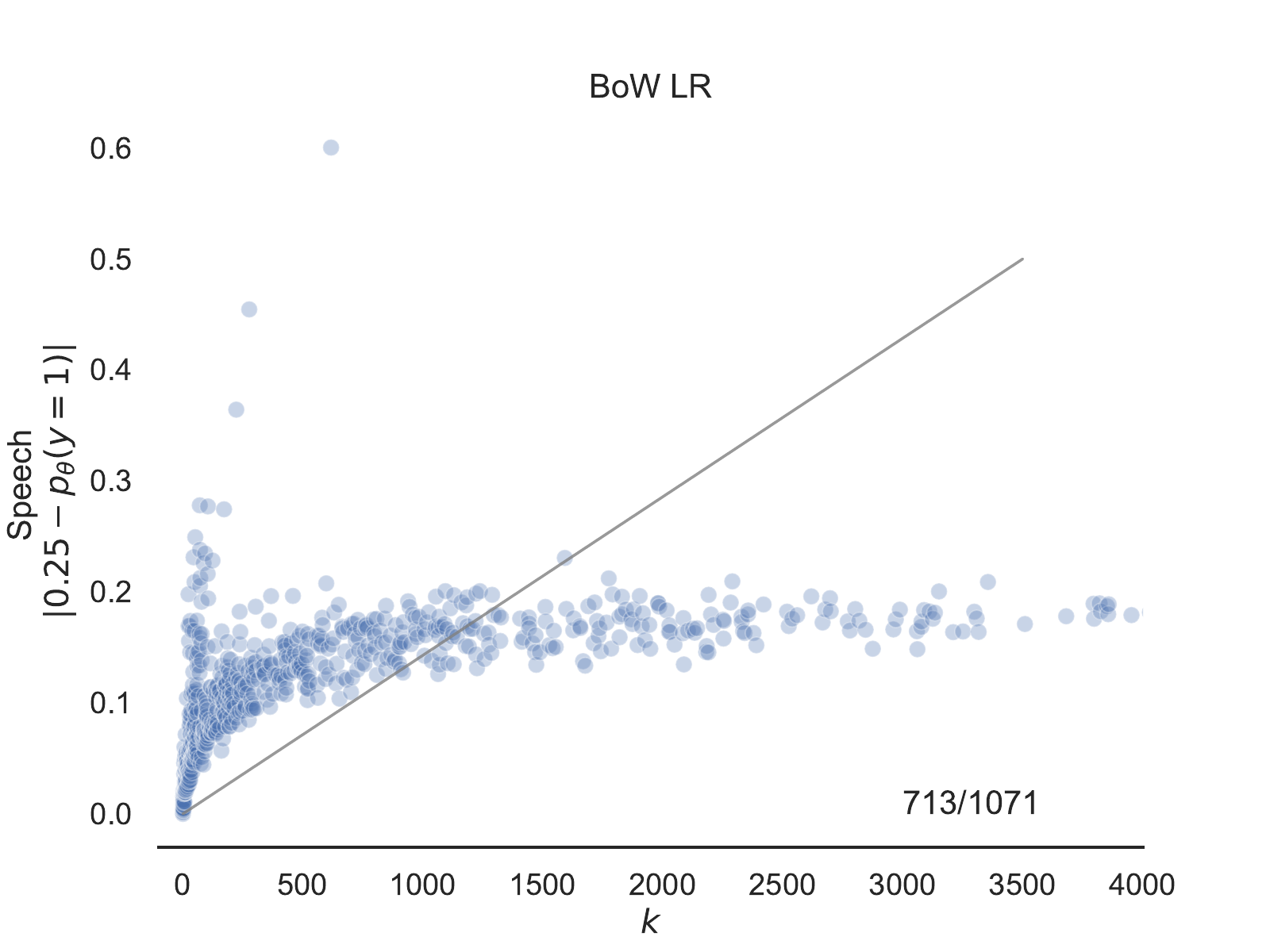}
\end{subfigure}
\hspace{1em}
\begin{subfigure}{.35\linewidth}
    \centering
    \includegraphics[width=\textwidth]{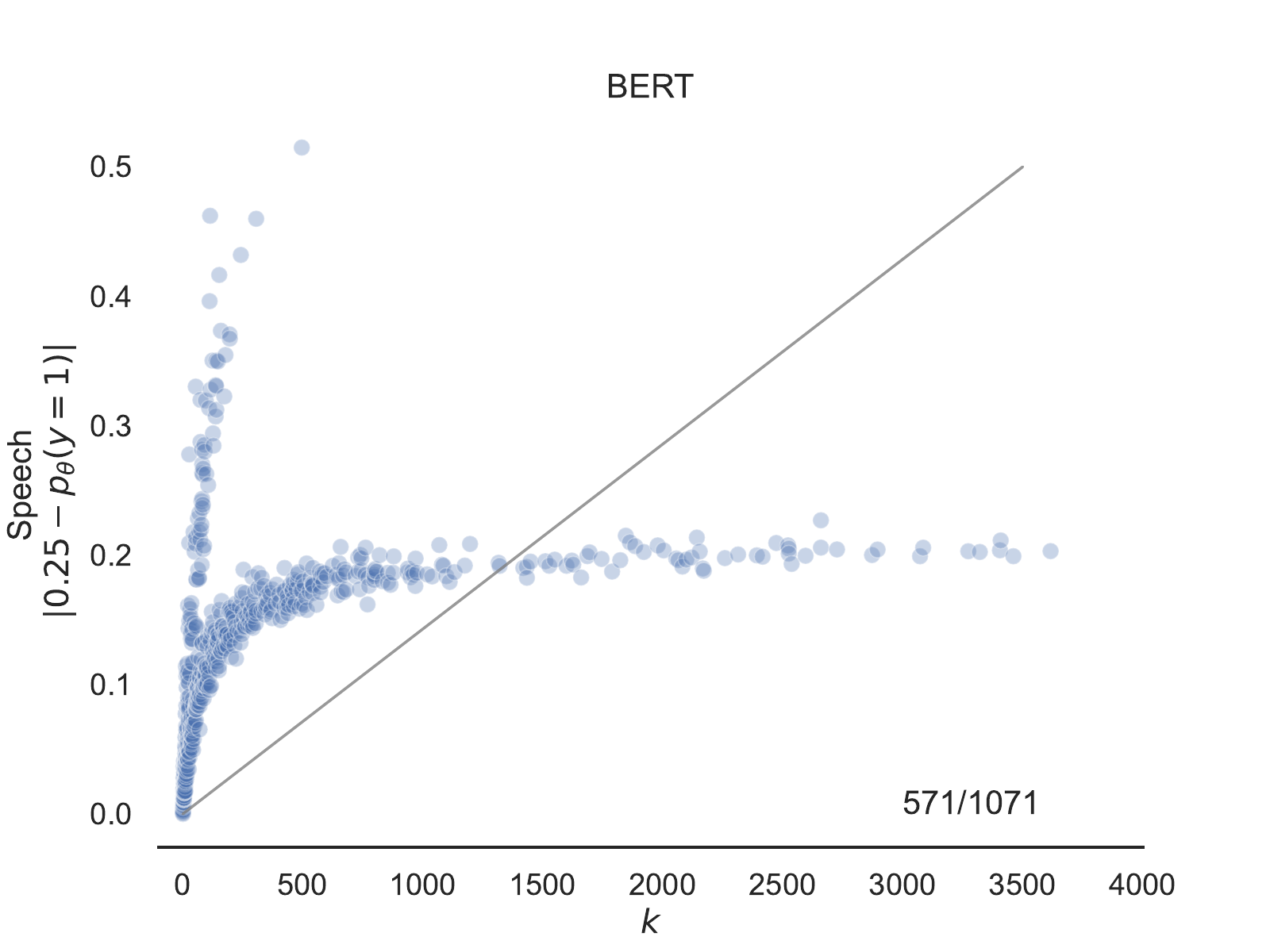}
\end{subfigure}

\hspace{1em}

\begin{subfigure}{.35\linewidth}
    \centering
    \includegraphics[width=\textwidth]{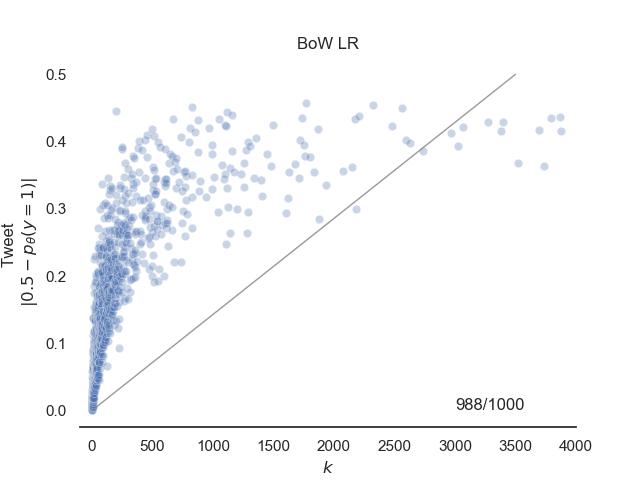}
\end{subfigure}
\hspace{1em}
\begin{subfigure}{.35\linewidth}
    \centering
    \includegraphics[width=\textwidth]{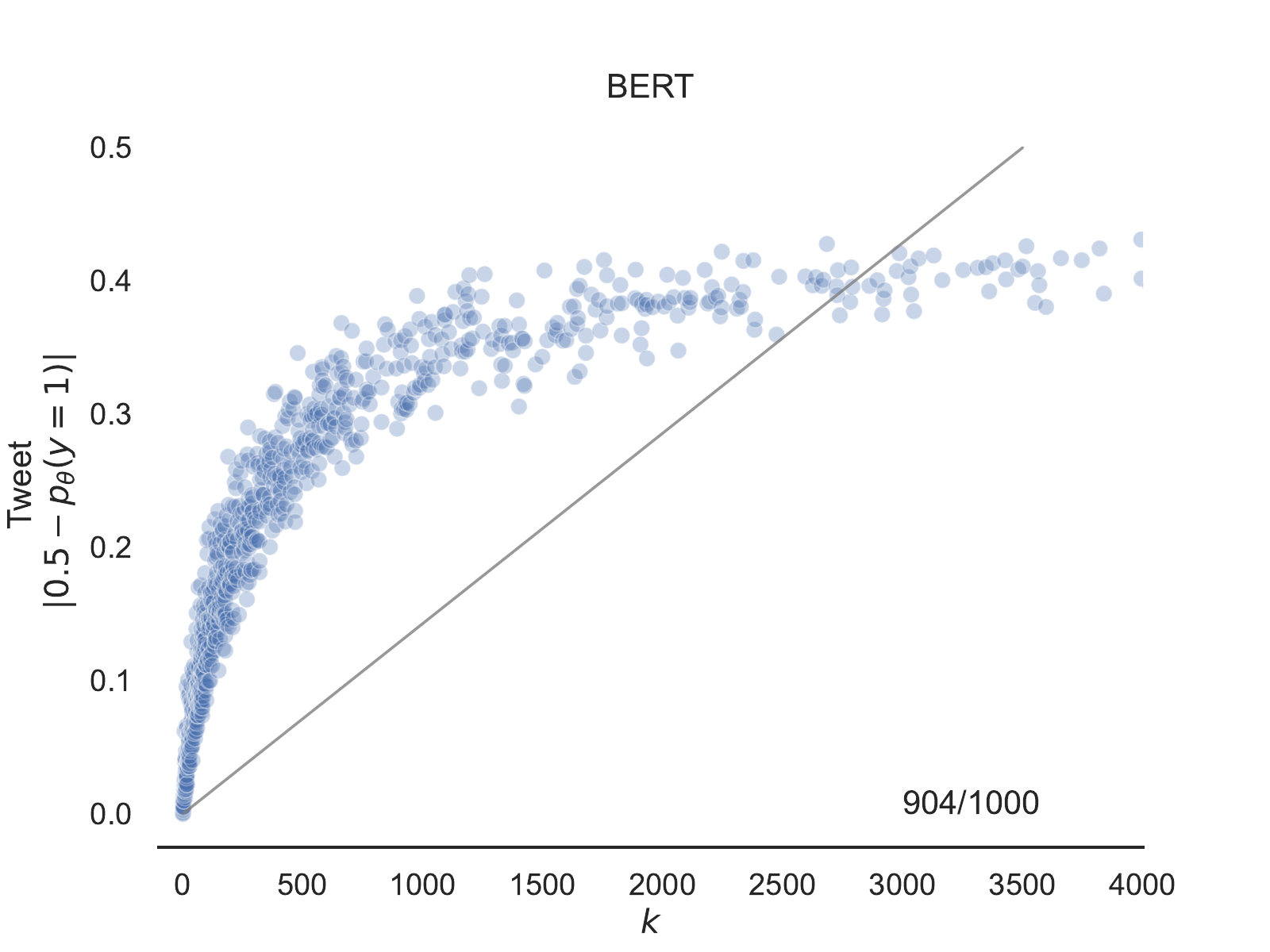}
\end{subfigure}

\caption{Relationship between predicted probabilities and  $k=|\mathcal{S}_t|$ identified from Algorithm \ref{alg:alg1}.}
\label{fig:kp_alg1}
\end{figure*}

\begin{figure*}
\centering
\begin{subfigure}{.425\linewidth}
    \centering
    \includegraphics[width=\textwidth]{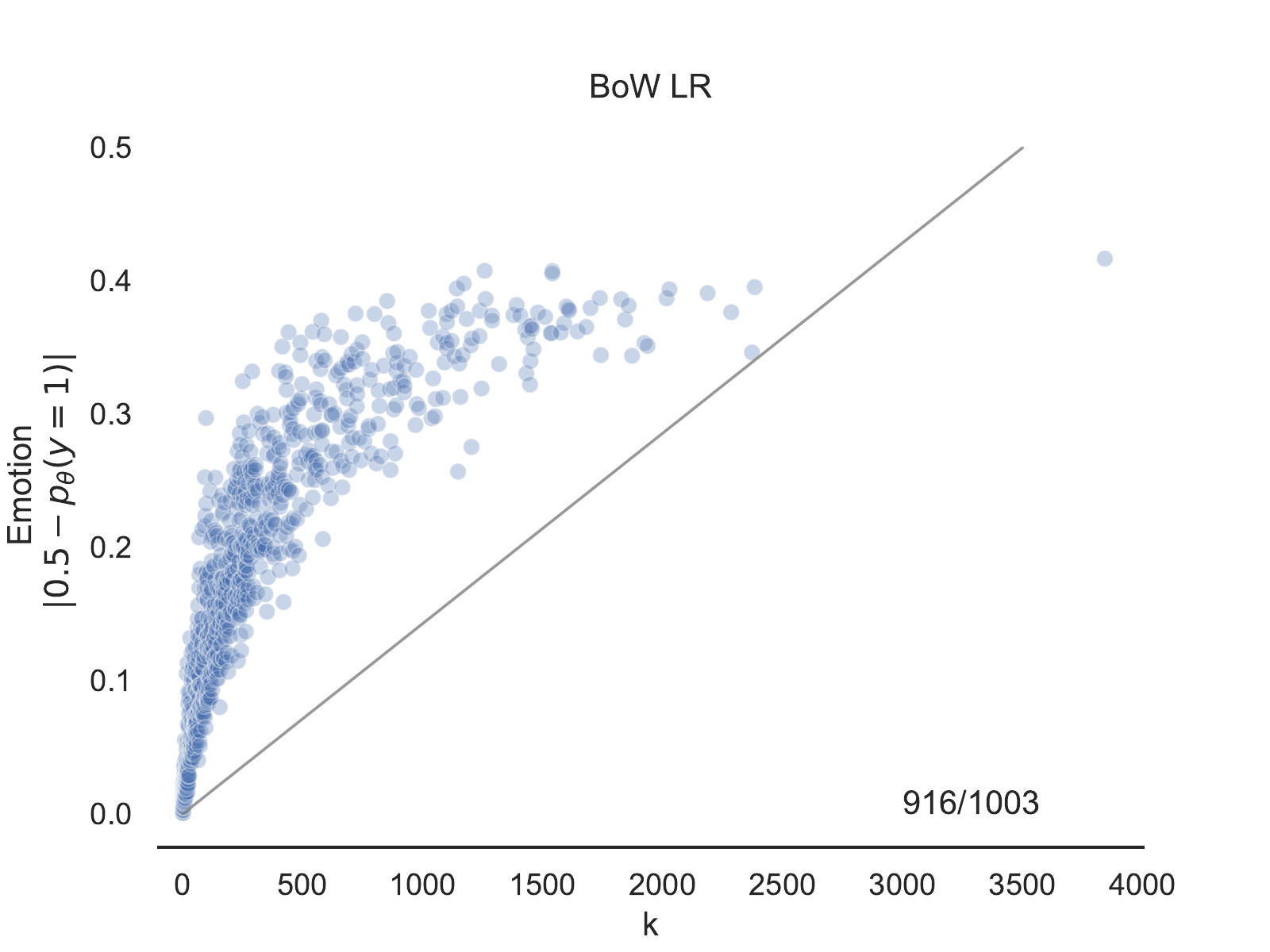}
\end{subfigure}
\hspace{1em}
\begin{subfigure}{.425\linewidth}
    \centering
    \includegraphics[width=\textwidth]{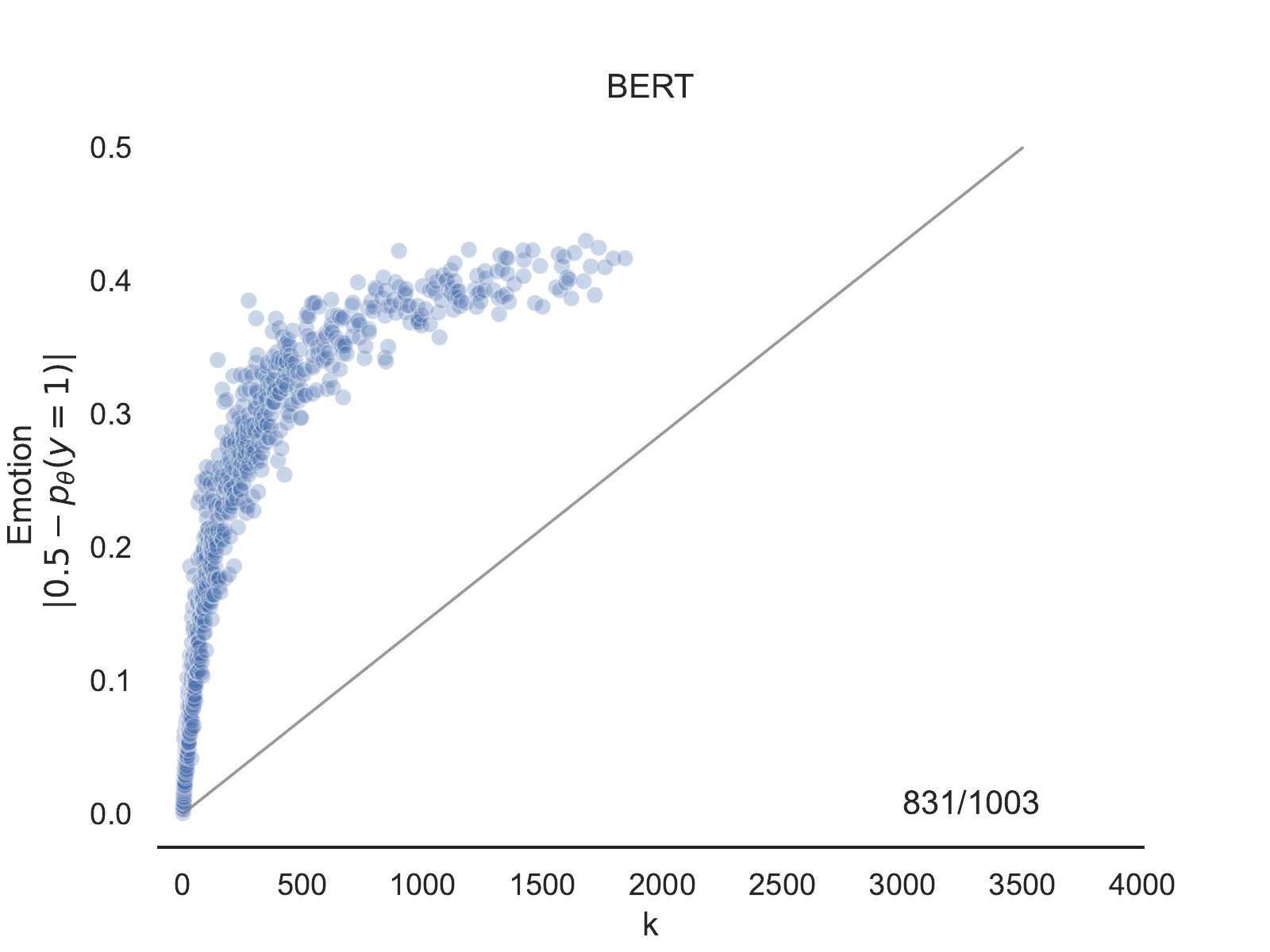}
\end{subfigure}

\hspace{1em}
\begin{subfigure}{.425\linewidth}
    \centering
    \includegraphics[width=\textwidth]{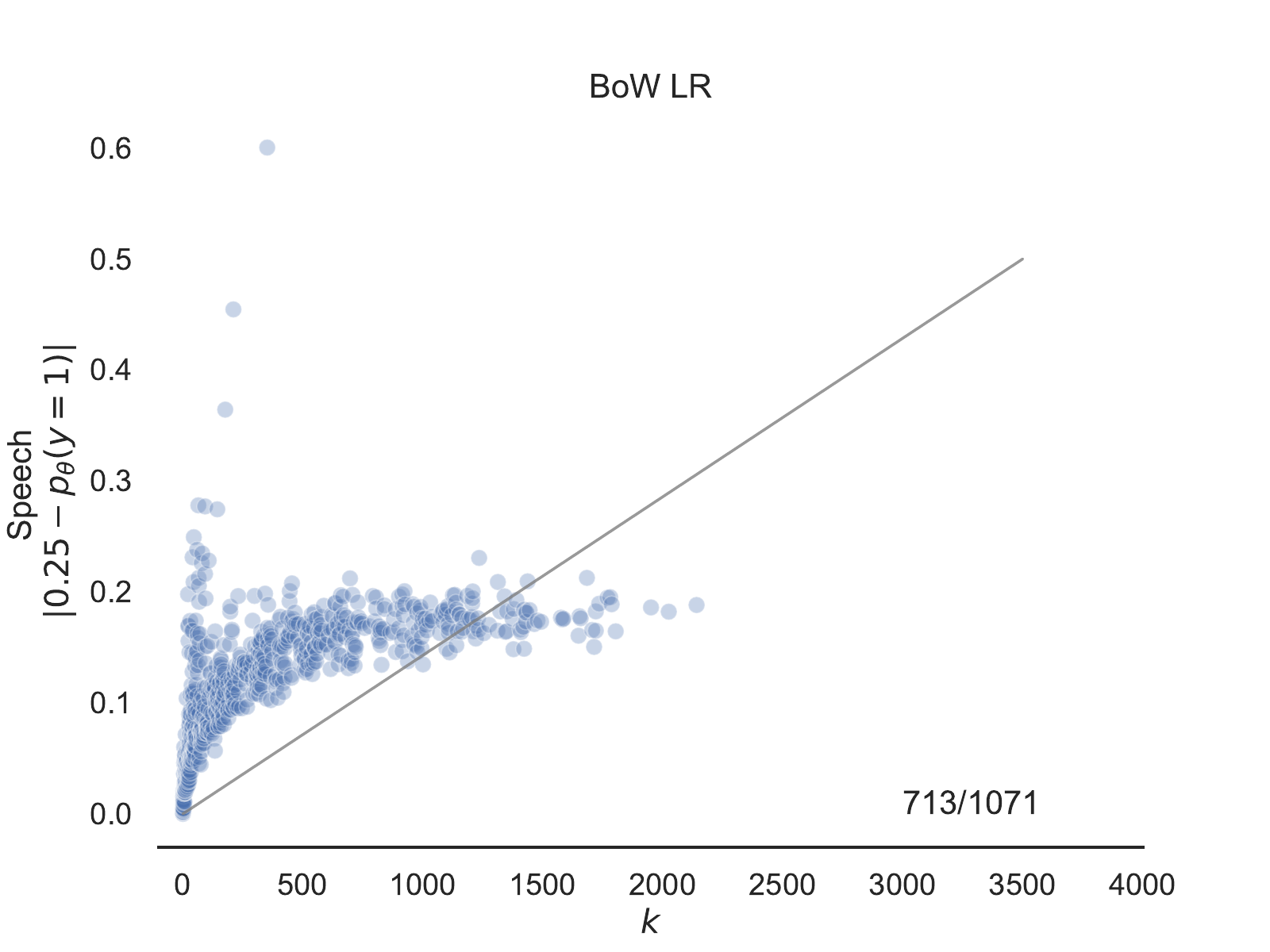}
\end{subfigure}
\hspace{1em}
\begin{subfigure}{.425\linewidth}
    \centering
    \includegraphics[width=\textwidth]{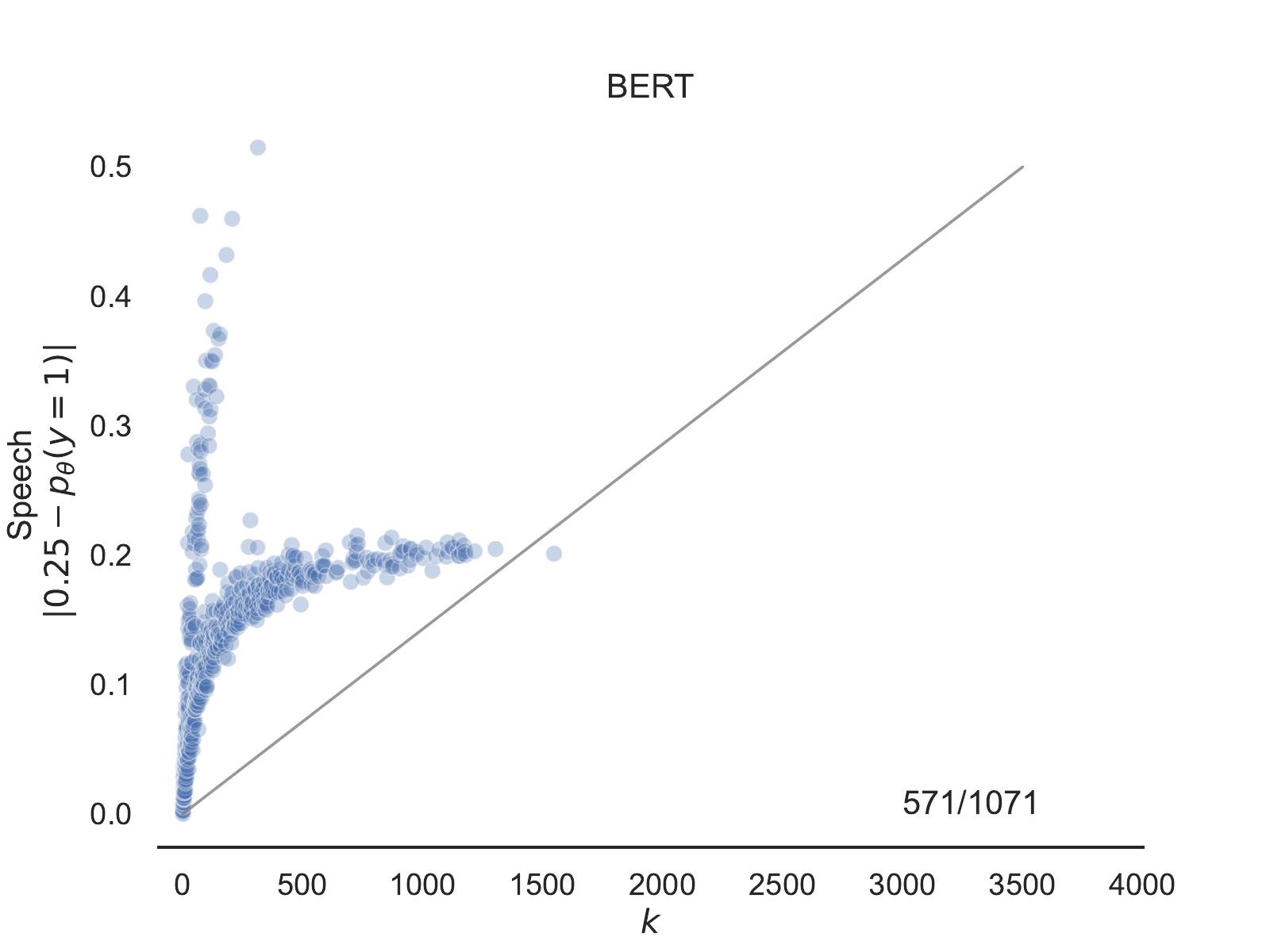}
\end{subfigure}

\hspace{1em}
\begin{subfigure}{.425\linewidth}
    \centering
    \includegraphics[width=\textwidth]{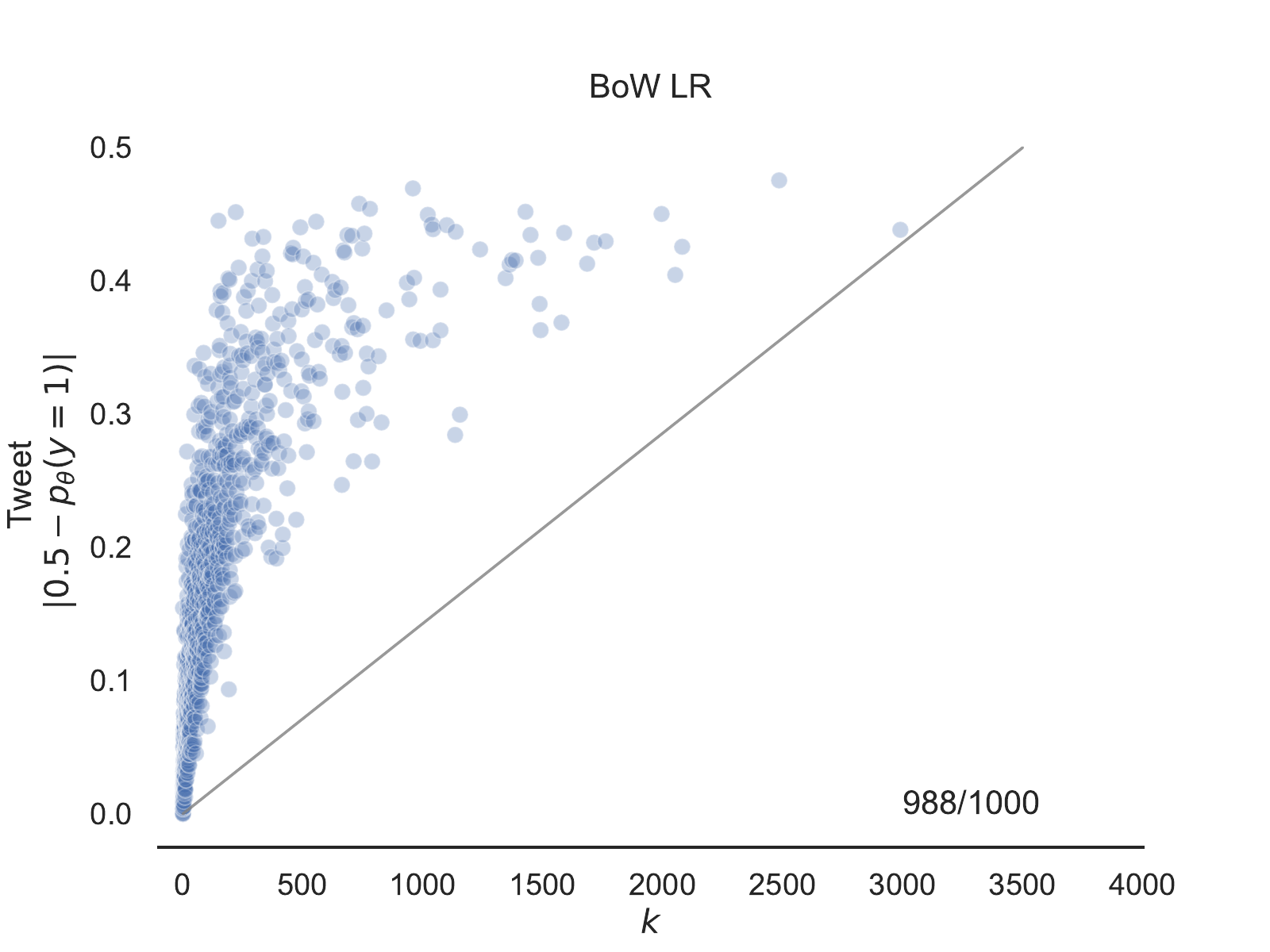}
\end{subfigure}
\hspace{1em}
\begin{subfigure}{.425\linewidth}
    \centering
    \includegraphics[width=\textwidth]{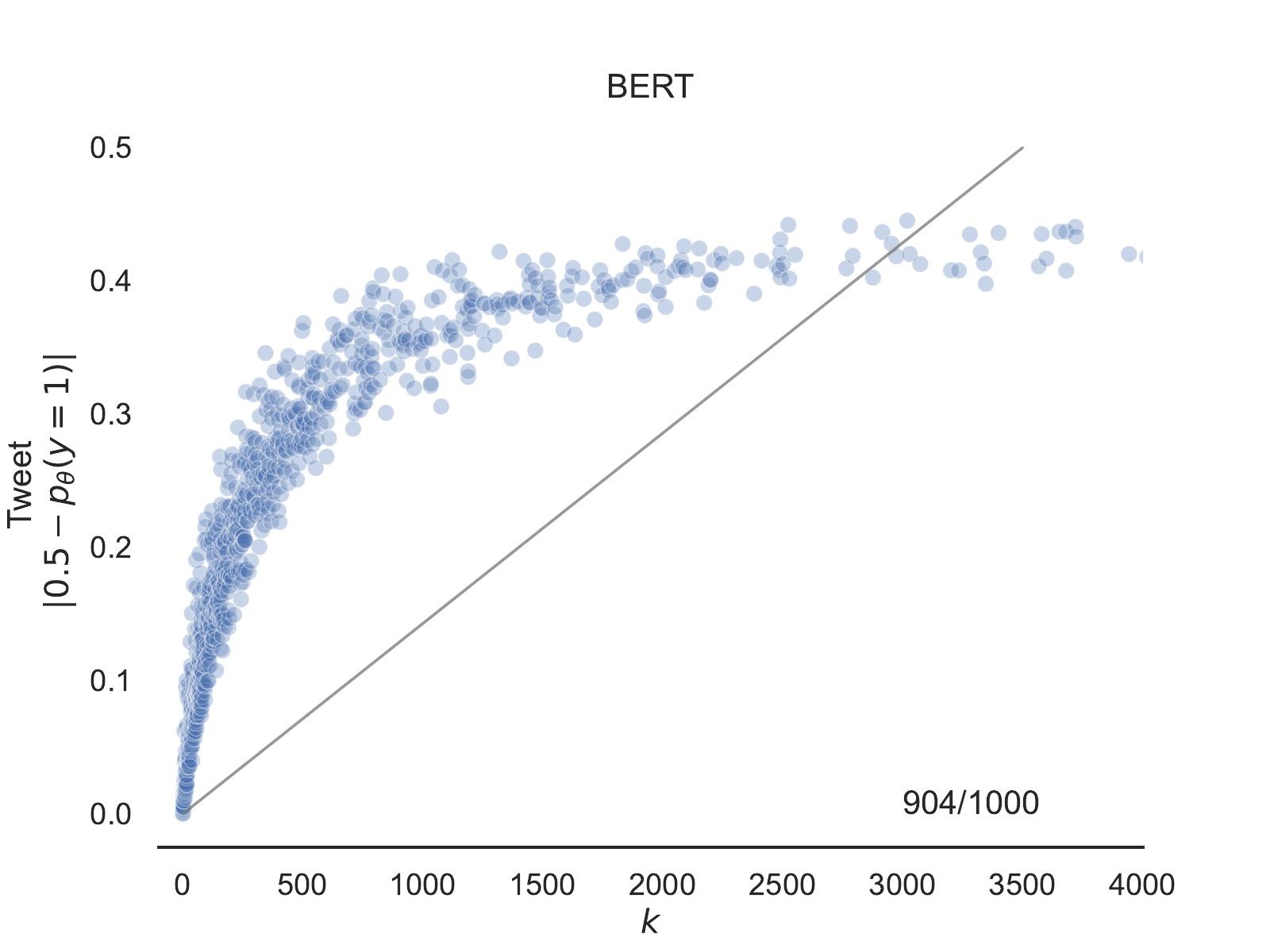}
\end{subfigure}

\caption{Relationship between predicted probabilities and  $k=|\mathcal{S}_t|$ identified from Algorithm \ref{alg:iterative}.}
\label{fig:kp_alg2}
\end{figure*}

Finally, we plot the relationship between $k$ and predicted probabilities under Algorithms \ref{alg:alg1} and \ref{alg:iterative} in Figures \ref{fig:kp_alg1} and \ref{fig:kp_alg2}, respectively.

\begin{table}
\small
\centering 
\begin{tabular}{l l l}
\hline
{\textbf{Features}} & {\textbf{Found $\mathcal{S}_t$}} & {\textbf{Flip successful}} \\
\hline 
 \multicolumn{3}{c}{\emph{Movie reviews}} \\
BoW & 78\% & 78\% \\
BERT & 79\% &  76\% \\
\multicolumn{3}{c}{\emph{Essays}} \\
BoW          &  12\% & 12\% \\ 
BERT &  9\% & 9\%\\
 \multicolumn{3}{c}{\emph{Emotion classification}} \\
BoW          & 91\% & 91\% \\ 
BERT & 83\% & 78\% \\
 \multicolumn{3}{c}{\emph{Hate speech}} \\
BoW          &  67\% & 65\%    \\ 
BERT &  53\% & 49\%     \\
\multicolumn{3}{c}{\emph{Tweet sentiment}} \\
BoW         & 99\% & 98\%  \\
BERT        & 90\% & 73\%   \\
\hline
\end{tabular}

\caption{Percentages of test examples for which Algorithm \ref{alg:alg1} successfully identified a set $\mathcal{S}_t$ (center) and for which upon removing these instances and retraining the prediction indeed flipped (right).}
\label{table:algo1-flips}
\end{table}
 
\subsection{Additional qualitative analysis}

We conclude with a brief qualitative analysis of examples in $\mathcal{S}_t$ retrieved in the case of the essays data. 
The model operating over BERT representations classified this test point ($x_t$) as 0, i.e., not an ``A'': ``\emph{The cyclist in this essay was a very brave man} ...''. The example is about a paragraph in length total, but details adventures of a cyclist. 
In this case it happens that the reference label is, in fact, an ``A'', so the model is incorrect.
Algorithm \ref{alg:iterative} reveals that removing a single training point and retraining would have overturned this prediction, yielding an ``A''.
The point in question is labeled 0 (so below an ``A'') and is about the mood of a memoir, in particular arguing that the person central to this was happy. 
The student-author of the cyclist essay might reasonably argue that this example is not at all relevant to their essay, and the fact that excluding this single example would have meant their essay received an ``A'' may be an adequate case for changing their grade accordingly.

\subsection{Time complexity}

We recorded wall clock times required to search for $|\mathcal{S}_t|$ on all test points in each dataset using Algorithm \ref{alg:alg1} and Algorithm \ref{alg:iterative} on Intel(R) Core(TM) i9-9920X CPUs; we report these times in Table \ref{table:running-time}. 
For Algorithm \ref{alg:alg1}, the longest running time is required for the essay dataset because most test predictions cannot be flipped even after iterating over all training points. 
Algorithm \ref{alg:iterative} is considerably slower than Algorithm \ref{alg:alg1}. 
The main reason lies in recording the set of training points not in $\mathcal{S}_t$ (line 20 in Algorithm \ref{alg:iterative}) and re-calculating the IP value in each iteration to reduce the minimal candidate set.
This additional time is traded off against the ability to (typically) find smaller $\mathcal{S}_t$ compared with \ref{alg:alg1}.
Overall, the running time required to find $|\mathcal{S}_t|$ for one test point is relatively minimal for both algorithms. 

\subsection{Attribution methods}

We consider different methods (i.e., other than influence functions) to rank training instances including gradient similarities in terms of the loss, similarity-based methods and randomly sampling training points. Of these we found that the proposed method works best in terms of finding instances which exert maximal influence on the prediction.

One natural way to quantify the impact of a training point $x_i$ on a training point $x_t$ by similarity methods. If the model has training points similar to the test point, it may classify the test correctly with high probability. We consider three of similarity-based methods: $\textbf{EUC} = -||x_t - x_i||^2$, $\textbf{DOT} = \left< x_i, x_t \right>$, and $\textbf{COS} = cos(x_i, x_t)$.

Apart from influence function and IP, we consider gradient-based instance attribution methods:

1) $RIF = cos(H^{-\frac{1}{2}} \nabla_{\theta}\mathcal{L}(x_t), H^{-\frac{1}{2}} \nabla_{\theta}L(x_i))$

2) $GD = \left< \nabla_\mathcal{L}L(x_t), \nabla_{\theta}L\mathcal{L} \right>$

3) $GC = cos( \nabla_{\theta}\mathcal{L}(x_t),  \nabla_{\theta}\mathcal{L}(x_i))$

 RIF was proposed to 
 mitigate the issues of outliers and mislabeled points being returned by the standard influence functions \cite{RIF}. GC and GD measure the similarity between two instances can also become an effective way to interpret the model from the instance perspective \cite{GC}. Apart from the methods above, we randomly sample training subsets and remove them accordingly.
 
\begin{figure}
    \centering
    \includegraphics[scale=0.33]{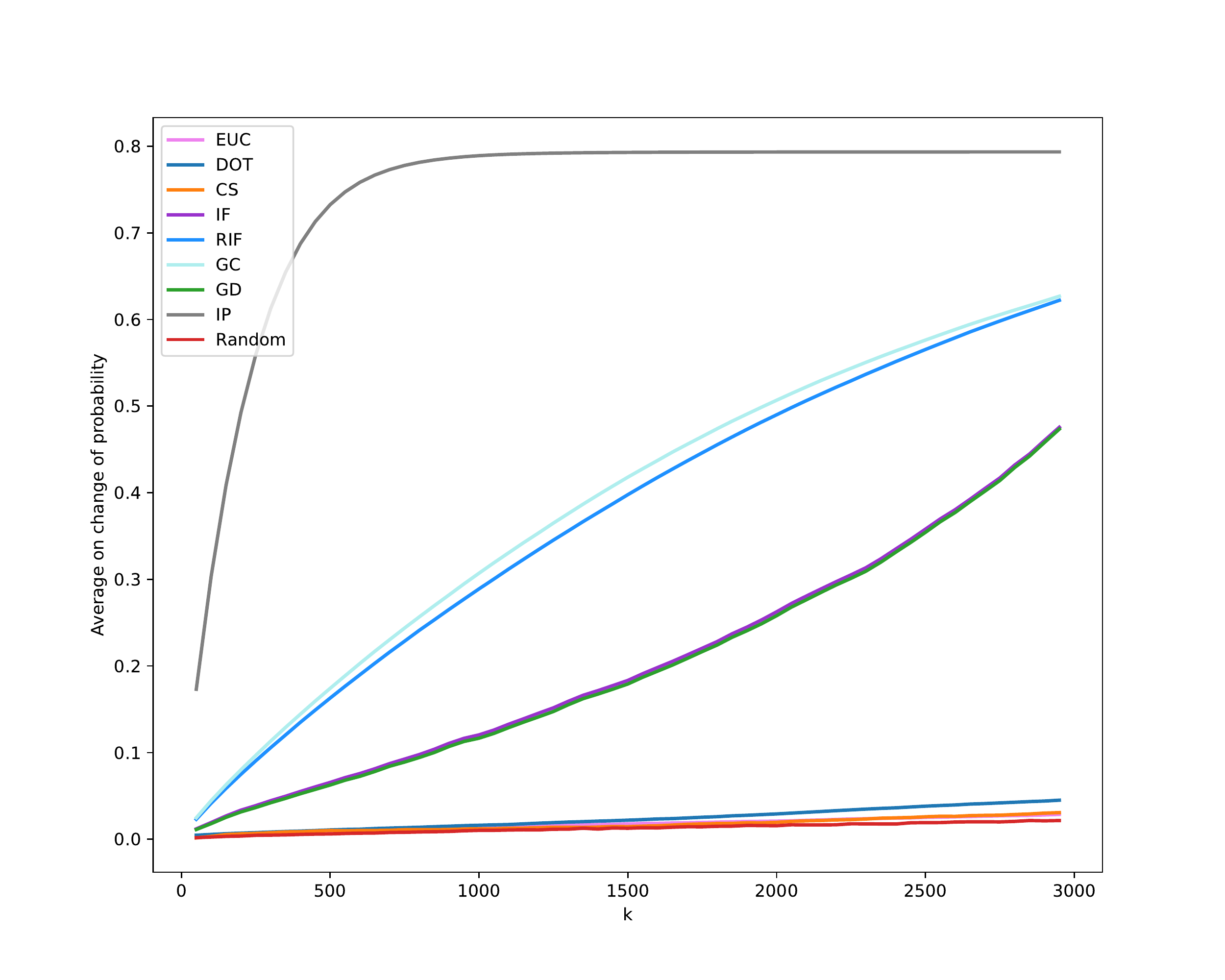}
    \caption{The relationship between the mean of absolute difference on predicted probabilities for all test points results from removing $|\mathcal{S}_t|$ training points, using different methods.}
    \label{fig:count}
\end{figure}

 We apply the above methods to the movie review dataset trained with a logistic regression model. We evaluate each attribution method as follows: First, we remove the top $k=|\mathcal{S}_t|$ training points from the training dataset according to the score calculated from the attribution method. Then we train a new with the same dataset except for the removed points. Finally, we compare the difference in predictions for each test point from the old model to the new model. To show the impact of attribution methods under different $k=|\mathcal{S}_t|$, we iterated with $k=|\mathcal{S}_t|$ from 50 to 3000. The mean absolute difference is plotted along with $k=|\mathcal{S}_t|$ shown in Figure \ref{fig:count}. 
 IP has a larger impact on the predicted probability, compared to removing training points ranked according to other methods.

 \begin{table*}
\small
\centering 
\begin{tabular}{l l l l l l l }
\hline
 \text{Datasets} & & \text{Movie reviews} & {\text{Essays}} & {\text{Emotion}} & {\text{Hate speech}} & {\text{Tweet}} \\
\hline 
{Bow} & Algorithm \ref{alg:alg1} & 5 & 155 & 5 & 40 & 11\\
& Algorithm \ref{alg:iterative} & 239 & 257 & 534 & 444 & 1529 \\
\hline 
{BERT} & Algorithm \ref{alg:alg1} & 3 & 161 & 19 & 52 & 8\\
& Algorithm \ref{alg:iterative} & 604 & 288 & 761 & 522 & 2203 \\
\hline
\end{tabular}
\caption{Running time (in seconds) to find $|\mathcal{S}_t|$ for all test points in each data set by Algorithm \ref{alg:alg1} and Algorithm \ref{alg:iterative}.}
\label{table:running-time}
\end{table*}

\begin{algorithm*}
\small
\caption{An \emph{iterative} approach to finding a minimal set to flip a prediction} 
\label{alg:iterative}
\DontPrintSemicolon
  \KwInput{$f$: Model;
  $Z^{\text{tr}}$: Full training set; 
  $\hat{\theta}$: Parameters estimated $Z^{\text{tr}}$;  
  $\mathcal{L}$: Loss function; 
  $x_{t}$: A test point; 
  $\tau$: Classification threshold (e.g., 0.5)}
    \KwOutput{$\mathcal{S}_t$: minimal train subset identified to flip prediction for $x_t$ ($\emptyset$ if unsuccessful)}
  $\theta' \leftarrow \hat{\theta}$ \\
  $Z^{\text{tr}}_{\text{r}}, \tilde{\mathcal{S}}_t 
  \leftarrow Z^{\text{tr}}, Z^{\text{tr}}$ \tcp{Track remaining points and candidate subset $\tilde{\mathcal{S}}_t$} 
  $H \leftarrow \nabla_{\theta}^2  \mathcal{L}(Z^{\text{tr}}, \theta')$ ~\\ 
  $\Delta \theta  \leftarrow H^{-1} \nabla_{\theta} \mathcal{L}(Z^{\text{tr}}, \theta')$ ~\\ 
  $\Delta_t f \leftarrow \nabla_{\theta} f_{\hat{\theta}}(x_{t})^\intercal \Delta \theta$ ~\\
  $\hat{y}_t \leftarrow f(x_{t}) > \tau$ ~\\
  $\Delta_t f_{\text{sum}} \leftarrow 0$ ~\\
  $k' \leftarrow |Z^{\text{tr}}|$ ~\\
    \While{$\mathcal{\tilde{S}}$ \text{changed since last iteration}}
    {
    \tcp{Sort instances (and estimated output differences) in order of the current prediction} 
    ${\tt direction} \leftarrow \{\uparrow$ if $\hat{y}_t$ else $\downarrow$\} ~\\
    ${\tt indices} \leftarrow {\tt argsort}(\Delta_t f, {\tt direction})$~\\
    $\Delta_t f \leftarrow {\text{\tt sort}}(\Delta_t f, {\tt direction})$ ~\\
        \vspace{0.3em}
        \For{$k=1$ ... $|\tilde{\mathcal{S}_t}|$}
   	    {
   	    $\hat{y}_t' \leftarrow (f(x_{t}) + {\text{\tt sum}}(\Delta_t f [:k])) > \tau$ ~\\
   	        \If{$\hat{y}'_t \neq \hat{y}_t$} 
            {   
                $\Delta_t f_{\text{sum}} \leftarrow {\text{\tt sum}}(\Delta_t f [:k])$ ~\\
                ${\tt diff} \leftarrow k' - k$~\\
                $k' \leftarrow k$ ~\\
                $\tilde{\mathcal{S}}_t \leftarrow  \mathcal{\tilde{S}}_t[{\tt indices}[:k]]$ \tcp*{Update candidate subset} 
                $Z^{\text{tr}}_{\text{r}} \leftarrow Z^{\text{tr}} / \mathcal{\tilde{S}}_t$ \tcp*{And the set of training points not in $\mathcal{\tilde{S}}_t$} 
                $\theta' \leftarrow  \theta + \Delta \theta[{\tt
                indices}[:k]] $ ~\\
                \tcp{Update Hessian and $\nabla$ of loss \emph{using updated $\theta$ estimate}}
                $H \leftarrow \nabla_{\theta}^2  \mathcal{L}(Z^{\text{tr}}_{\text{r}}, \theta')$  ~\\
                $\Delta \theta \leftarrow H^{-1} \nabla_{\theta} \mathcal{L}(\mathcal{\tilde{S}}_t, \theta')$ ~\\
                $\Delta_t f \leftarrow \nabla_{\theta} f_{\hat{\theta}}(x_{t})^\intercal \Delta \theta$~\\
                ${\tt break}$~\\
            }
   	    }
    }
        \If{$|\tilde{\mathcal{S}}_t| = |Z^{tr}|$}
        {
            \KwRet {$\emptyset$}
        }        
    \KwRet $\tilde{S}_t$
\end{algorithm*}

\end{document}